\pdfoutput=1

\documentclass[11pt]{article}

\usepackage{ACL2023}

\usepackage{hyperref}

\usepackage{times}
\usepackage{latexsym}

\usepackage[T1]{fontenc}

\usepackage[utf8]{inputenc}

\usepackage{microtype}

\usepackage{inconsolata}

\usepackage{graphicx}
\usepackage[export]{adjustbox}
\usepackage{caption}
\usepackage[list=false]{subcaption}
\usepackage{comment}
\usepackage[linesnumbered,ruled,vlined]{algorithm2e}

\SetCommentSty{mycommfont}

\usepackage{amsfonts}
\usepackage{mathtools, cases}

\usepackage{footnote}

\usepackage{tablefootnote}
\usepackage{tabularx}
\usepackage{multicol}
\usepackage{multirow}
\usepackage{tablefootnote}
\usepackage{color, colortbl}
\usepackage{algpseudocode}
\definecolor{Gray}{gray}{0.9}

\usepackage{wasysym} 
\usepackage{tcolorbox}

\usepackage{amssymb}

\usepackage{pifont}

\usepackage{calc}

\newlength\WIDTHOFBAR
\usepackage{tikz}
\usepackage{verbatim}
\usetikzlibrary{arrows,positioning,shapes.geometric,decorations.pathreplacing, calc, quotes, arrows.meta}

\usepackage{enumitem}

\makeatletter
\newcommand*{\radiobutton}{%
  \@ifstar{\@radiobutton0}{\@radiobutton1}%
}
\newcommand*{\@radiobutton}[1]{%
  \begin{tikzpicture}
    \pgfmathsetlengthmacro\radius{height("X")/2}
    \draw[radius=\radius] circle;
    \ifcase#1 \fill[radius=.6*\radius] circle;\fi
  \end{tikzpicture}%
}
\makeatother

\definecolor{cambridgeblue}{rgb}{0.64, 0.76, 0.68}
\definecolor{blue(ncs)}{rgb}{0.0, 0.53, 0.74}
\definecolor{cadetblue}{rgb}{0.37, 0.62, 0.63}
\definecolor{cadmiumgreen}{rgb}{0.0, 0.42, 0.24}
\definecolor{ao(english)}{rgb}{0.0, 0.5, 0.0}
\definecolor{alizarin}{rgb}{0.82, 0.1, 0.26}
\definecolor{vermilion}{rgb}{0.89, 0.26, 0.2}

\makeatletter
\def\thickhline{%
  \noalign{\ifnum0=`}\fi\hrule \@height \thickarrayrulewidth \futurelet
   \reserved@a\@xthickhline}
\def\@xthickhline{\ifx\reserved@a\thickhline
               \vskip\doublerulesep
               \vskip-\thickarrayrulewidth
             \fi
      \ifnum0=`{\fi}}
\makeatother

\newlength{\thickarrayrulewidth}
\title{NLG Evaluation Metrics Beyond Correlation Analysis: \\An Empirical Metric Preference Checklist}

\author{Iftitahu Ni'mah\textsuperscript{\ding{168},\ding{171}} \hspace{1.5mm} Meng Fang\textsuperscript{\ding{169}} \hspace{1.5mm} Vlado Menkovski\textsuperscript{\ding{168}} \hspace{1.5mm} Mykola Pechenizkiy\textsuperscript{\ding{168}}    \\
{\normalsize{\textsuperscript{\ding{168}} Eindhoven University of Technology \hspace{1.5mm}\textsuperscript{\ding{169}} University of Liverpool \hspace{1.5mm}\textsuperscript{\ding{171}} BRIN Indonesia}} \\
\texttt{\small{\{i.nimah, v.menkovski, m.pechenizkiy\}@tue.nl, Meng.Fang@liverpool.ac.uk}}}

\begin{document}
\maketitle
\begin{abstract}

In this study, we analyze automatic evaluation metrics for Natural Language Generation (NLG), specifically task-agnostic metrics and human-aligned metrics. Task-agnostic metrics, such as Perplexity, BLEU, BERTScore, are cost-effective and highly adaptable to diverse NLG tasks, yet they have a weak correlation with human. Human-aligned metrics (CTC, CtrlEval, UniEval) improves correlation level by incorporating desirable human-like qualities as training objective. However, their effectiveness at discerning system-level performance and quality of system outputs remain unclear.

We present metric preference checklist as a framework to assess the effectiveness of automatic metrics in three NLG tasks: Text Summarization, Dialogue Response Generation, and Controlled Generation. Our proposed framework provides access: (i) for verifying whether automatic metrics are faithful to human preference, regardless of their correlation level to human; and (ii) for inspecting the strengths and limitations of NLG systems via pairwise evaluation. We show that automatic metrics provide a better guidance than human on discriminating system-level performance in Text Summarization and Controlled Generation tasks. We also show that multi-aspect human-aligned metric (UniEval) is not necessarily dominant over single-aspect human-aligned metrics (CTC, CtrlEval) and task-agnostic metrics (BLEU, BERTScore), particularly in Controlled Generation tasks. \footnote{Our code is available at \href{https://github.com/inimah/metric-preference-checklist}{https://github.com/inimah/metric-preference-checklist}.} 



\end{abstract}

\section{Introduction}

Natural Language Generation (NLG) refers to an automatic process to generate texts in one or more language categories that satisfy multiple desirable human-like qualities. For example, in Text Summarization \cite{novikova-etal-2017-need,maynez-etal-2020-faithfulness,bhandari-etal-2020-evaluating,fabbri-etal-2021-summeval}, NLG systems are expected to produce \emph{coherent, consistent, fluent, }and \emph{relevant} summarization outputs. In Dialogue Response Generation \cite{see-etal-2019-makes}, the system outputs are mainly assessed based on aspects that are important in a typical human conversation, such as \emph{naturalness} and \emph{engagingness}. In Controlled Generation \cite{Dathathri2020Plug}, the generation outputs are evaluated based on its \emph{relevance} to the predefined topic category or sentiment category as control attributes.

A standard evaluation protocol in NLG for assessing the above human-like qualities involves conducting a human evaluation study or running an automatic evaluation, or both ways. A human evaluation study improves the reliability of evaluation process, particularly when the assessment is done by experts. It is also often infeasible to translate human evaluation aspects into an automatic statistical metric formulation due to its multi-dimensional abstractive properties \cite{birch-etal-2013-feasibility,hashimoto-etal-2019-unifying}. However, human evaluation is known to be more costly and does not scale well \cite{howcroft-etal-2020-twenty,10.1162/tacl_a_00437}. Utilizing automatic metrics, on the other hand, is cost-effective and more feasible for large-scale evaluation data.

\begin{table*}[!t]
    \centering
    \resizebox{.995\textwidth}{!}{
    \begin{tabular}{p{3.8cm}  p{5.75cm}  p{5.5cm} }
     \hline
     \thickhline
      \bf Assessment Type  & \bf Description & \bf Research Question  \\
     \hline
     \thickhline
    \hline
    Transfer experiment & Correlation analysis between automatic metrics and human judgments in In-Domain (ID) and Out-of-Domain (OOD) use cases. & Is correlation level to human judgments consistent across ID and OOD use cases? \\
    Aspect-level evaluation & Evaluating metric's effectiveness at identifying different levels of human-like quality.  &  Is human-aligned metric better at distinguishing between different levels of human-like quality of system outputs?   \\
    Aspect-level preference &  Preference similarity between human and automatic metrics on identifying different levels of human-like quality &  Do human and automatic metrics rank the quality of system outputs similarly?   \\
    System-level evaluation & Evaluating the metric effectiveness at discerning system-level performance & Is human-aligned metric better at discerning performance of independent NLG systems?  \\
    System-level preference &  Preference similarity between human and automatic metrics on identifying the performance rank of the systems. &  Do human and automatic metrics rank systems similarly?   \\

    \hline
    \thickhline
    \end{tabular}
    }
    \caption{Metric preference checklist. }
    \label{tab:preference-check}
\end{table*}

Recent works on automatic NLG evaluation metrics, such as CTRLEval \cite{ke-etal-2022-ctrleval}, CTC \cite{deng-etal-2021-compression}, and UniEval \cite{zhong-etal-2022-towards}, have made progress in improving the correlation between automatic metrics and human up to 43\% by developing human-aligned automatic metrics. Despite the advancements, there is a need for a standardized framework to assess the utility of these metrics in the context of discerning system-level performance. The reason is that an overall correlation score to human does not necessarily represents the metric effectiveness as an evaluation tool, as demonstrated by previous analysis studies on NLG automatic metrics \cite{caglayan-etal-2020-curious,hanna-bojar-2021-fine,sai-etal-2021-perturbation,10.1145/3485766}. However, none of these works connect the correlation analysis to the metric effectiveness at addressing the main objective of NLG benchmarking. That is, for distinguishing system-level performance.






Our study addresses the above research gap by designing a metric preference checklist for measuring the effectiveness of automatic metrics in three NLG tasks: Text Summarization (TextSumm), Dialogue Response Generation (DiagGen), and Controlled Generation (CtrlGen). We introduce three types of assessment for evaluating NLG automatic metrics: Transfer experiment, Aspect-level evaluation, and System-level evaluation. The implications of our study are threefold:

\begin{itemize}
    \item Verifying the faithfulness of automatic metrics to human preference is a necessary component for a more accurate interpretation of evaluation outcomes (section \textsection{\ref{sec:faithful}}).


    \item Automatic metrics can be more discriminating than human (section \textsection{\ref{sec:discriminate}}).

    \item Benchmarking NLG systems via pairwise comparison provides more insights into the strengths and limitations of the systems w.r.t. desirable human-like qualities (section \textsection{\ref{sec:benchmark}}).

\end{itemize}

\section{Related Work}
\label{sec:related-work}

Existing automatic metrics in NLG are mainly dominated by task-agnostic metrics -- metrics that assess the quality of generation outputs without considering human evaluation aspects as context or objective of the evaluation task \cite{10.1145/3485766}. Task-agnostic metrics are highly adaptable across NLG tasks because the adaptation does not require task-specific design. For example, BLEU \cite{papineni-etal-2002-bleu} and ROUGE \cite{lin2004rouge}, which represent string-based metrics, are largely adopted in Neural Machine Translation (NMT) and Text Summarization. Perplexity \cite{jelinek1977perplexity,brown-etal-1992-estimate} -- a reference-less metric, is a standard evaluation metric in a Language Modeling-based NLG tasks, including Controlled Generation \cite{Keskar2019CTRLAC,Dathathri2020Plug}. BERTScore \cite{Zhang*2020BERTScore:} has been largely adopted in diverse NLG tasks, including NMT \cite{colombo2022infolm}, Text Summarization \cite{deutsch-roth-2021-understanding}, and Dialogue System \cite{yeh-etal-2021-comprehensive}. Nevertheless, currently available task-agnostic metrics are weakly correlated to human judgment \cite{novikova-etal-2017-need,sai-etal-2021-perturbation,10.1145/3485766}. A low correlation score introduces a criticism on the capability of automatic metrics at identifying the different quality of system outputs and their potential usage to substitute a costly human evaluation study. 

Recent works \cite{deng-etal-2021-compression,ke-etal-2022-ctrleval,zhong-etal-2022-towards} have demonstrated that incorporating desirable human-like qualities as a training objective or contextual knowledge is the best-fit solution for improving the correlation level between automatic metrics and human. However, verifying whether a higher correlation represents a higher human preference for ranking the quality of system outputs and ranking system performance, and vice versa, remains an underexplored query. Compared to the recent analysis studies that focus on validating the robustness \cite{caglayan-etal-2020-curious,hanna-bojar-2021-fine,chen-etal-2021-factuality-checkers,vu-etal-2022-layer}, explainability \cite{kaster-etal-2021-global}, reproducibility \cite{chen-etal-2022-reproducibility}, and fairness \cite{sun-etal-2022-bertscore} of the NLG evaluation metrics, our study focuses more on a systematic assessment by connecting the link between correlation score to the practical use of the metrics in NLG evaluation. That is, (i) for discriminating the system outputs based on desirable human-like qualities; and (ii) for ranking system performance. 



\section{Metric Preference Checklist}
\label{sec:framework}

A standard evaluation protocol in NLG involves validating automatic metrics based on their correlation to human. Intuitively, a high correlation suggests a high agreement on discerning the quality of system outputs because low-quality outputs are penalized with lower scores, while high-quality outputs are rewarded with higher scores. However, currently available metrics are known to have a poor correlation to human. So, it is unclear to what extend current automatic metrics are capable of (i) identifying human-like quality of system outputs and (ii) discriminating performance between independent NLG systems.



To further investigate the above questions, we pose several relevant research questions as a metric preference checklist, as presented in Table~\ref{tab:preference-check}. We define the assessment tasks for evaluating NLG automatic metrics into five (5) fine-grained aspects, as follows:

\subsection{Transfer Experiment (Zero-Shot)} 
\label{sec:transfer}
The assessment is designed to investigate whether the correlations between automatic metrics and human are consistent across NLG use cases. For measuring the adaptability of automatic metrics in new target domain, we define In-Domain (ID) and Out-of-Domain (OOD) use cases as follows \footnote{We follow the categorization of OOD that is discussed in previous work by \citet{arora-etal-2021-types}.}:

\paragraph{In-Domain (ID)} For learnable or tunable automatic metrics, we define ID data as the dataset in which the metrics are introduced. For example, UniEval \cite{zhong-etal-2022-towards} is introduced with a subset of data from SummEval \cite{fabbri-etal-2021-summeval} and Topical-Chat \cite{mehri-eskenazi-2020-usr}. For task-agnostic metrics, such as Perplexity, BLEU, ROUGE, and BERTScore; the categorization of ID and OOD data is rather unknown. So, we define ID data based on a common sense perspective on how close a domain is to the NLG domain where the metric is introduced. For example, BLEU is originally introduced for a Neural Machine Translation (NMT) task \cite{papineni-etal-2002-bleu}, yet the metric is widely adopted in Text Summarization (TextSumm). Thus, datasets in Text Summarization domain are considered to be ID samples for BLEU metric.


\paragraph{Semantic-Shift OOD} Samples are drawn from the same domain or NLG task where the metric is introduced, but they do not necessarily contain overlapped semantic features with ID samples. For example, let consider ID samples $\{x, y\}$ are drawn from a subset of SummEval and Topical-Chat datasets introduced in UniEval benchmarking \cite{zhong-etal-2022-towards}. Semantic-Shift OOD samples are the superset $\{X, Y\}$, which are drawn from the original benchmark datasets of SummEval by \citet{fabbri-etal-2021-summeval} and Topical-Chat by \citet{mehri-eskenazi-2020-usr}. 


\paragraph{Domain-Shift OOD} Samples are drawn from a new domain where the human evaluation aspects overlap with ID domain, but the background features are different. For example, CTRLEval \cite{ke-etal-2022-ctrleval} is firstly introduced and evaluated in a Controlled Generation task. Thus, samples from different NLG use cases, such as Text Summarization and Dialogue Response Generation are considered to be a Domain-Shift OOD samples.

\begin{table*}[!ht]
    \centering
    \resizebox{.995\textwidth}{!}{
    \begin{tabular}{p{2.25cm} p{4.5cm} p{3.cm} c  p{4.5cm}}
     \hline
     \thickhline
    \bf NLG Task &  \bf Benchmark & \bf Data Abbreviation & \bf \#Samples & \bf Human-like Aspects \\
     \hline
     \thickhline
	& UBER-PPLM  \cite{Dathathri2020Plug}  &  UBER-Topic & 14626 & Fluency, Relevance \\
	CtrlGen  & CTRL  \cite{Keskar2019CTRLAC}  & CTRL-Topic & 3120 & Fluency, Relevance  \\
	& CTRL-Eval UBER \cite{ke-etal-2022-ctrleval} & CtrlEval-Topic & 960 & Coherence, Consistency, Fluency, Relevance \\
	\hline
	  & USR Persona chat \cite{mehri-eskenazi-2020-usr} & USR-PC & 900 & Understandable, Natural, MaintainsContext, Engaging, UsesKnowledge, Overall \\
	DiagGen & USR Topical chat \cite{mehri-eskenazi-2020-usr} & USR-TC & 1080 & Understandable, Natural, MaintainsContext, Engaging, UsesKnowledge, Overall \\
	& UniEval Topical chat \cite{zhong-etal-2022-towards} & UniEval-TC & 360 & Understandability, Naturalness, Coherence, Engagingness, Groundedness, Overall \\
	\hline 
	 & SummEval \cite{fabbri-etal-2021-summeval} & summEval & 5100 & Coherence, Consistency, Fluency, Relevance, Overall\\
    TextSumm	& Newsroom  \cite{grusky-etal-2018-newsroom} & Newsroom & 1260 & Coherence, Informativeness, Fluency, Relevance, Overall\\
	 & UniEval SummEval \cite{zhong-etal-2022-towards} & Unieval-summ & 1600 & Coherence, Consistency, Fluency, Relevance, Overall \\
	\hline
    \thickhline
    \end{tabular}
    }
    \caption{Benchmark datasets in this study. }
    \label{tab:humeval-data}
\end{table*}

\begin{table*}[!ht]
    \centering
    \resizebox{.995\textwidth}{!}{
    \begin{tabular}{p{3.5cm} p{3.cm}  p{3.2cm} p{3.2cm} p{3.2cm}  c}
     \hline
     \thickhline
      \bf Category  & \bf Metric & \bf ID  & \bf Semantic-Shift  & \bf Domain-Shift  &  \bf Human-aligned \\
     \hline
     \thickhline
       Surface-level   & \textbf{BLEU} & UniEval-summ, summEval, Newsroom & UniEval-TC, USR-TC, USR-PC & -& -\\
        & \textbf{ROUGE} & UniEval-summ, summEval, Newsroom   & UniEval-TC, USR-TC, USR-PC & - & - \\
       \hline
       Semantic similarity   & \textbf{BERTScore}  & UniEval-summ, summEval, Newsroom & UniEval-TC, USR-TC, USR-PC & UBER-Topic, CtrlEval-Topic& -\\
       \hline
       Language Model  & \textbf{Perplexity}   & UniEval-TC, USR-TC, USR-PC & UBER-Topic, CtrlEval-Topic & UniEval-summ, summEval, Newsroom & - \\
       \hline
	Information alignment & \textbf{CTC} \cite{deng-etal-2021-compression}  & CTC-TC, summEval, Newsroom & USR-TC, USR-PC & UBER-Topic, CtrlEval-Topic & \checkmark \\
	\hline
       Text Infilling & \textbf{CTRLEval} \cite{ke-etal-2022-ctrleval}  & CtrlEval-Topic & UBER-Topic, summEval, Newsroom  & USR-TC, USR-PC & \checkmark \\
       \hline
        Boolean QA & \textbf{UniEval} \cite{zhong-etal-2022-towards}  & UniEval-summ, UniEval-TC & summEval, Newsroom, USR-TC, USR-PC  & UBER-Topic, CtrlEval-Topic & \checkmark \\
    \hline
    \thickhline
    \end{tabular}
    }
    \caption{Automatic metrics and the corresponding datasets for transfer experiment. }
    \label{tab:stat-met}
\end{table*}

\subsection{System-level Evaluation}
The task's objective is to evaluate the effectiveness of the evaluation metrics at discerning the performance difference between independent NLG systems. For quantifying the degree to which the scores produced by automatic metrics are able to discern the performance between two different NLG systems, we utilize \textbf{Kolmogorov-Smirnov (KS)} as a statistical distance metric $D$:
\vspace{-.5em}
\begin{equation}
    \begin{aligned}
    D (P_1, P_2) = \sup_{s} |P_1(s) - P_2(s)|,
    \end{aligned}
\vspace{-.5em}
\label{eq:ks}
\end{equation}

\noindent where $P_1$ and $P_2$ denote the empirical cumulative density function (cdfs) of scores based on metric $M$ for system $A$ and system $B$, where $D \in [0,1] $. $s$ denotes the evaluation scores as random variables of metric $M$. $D (.)=0$ indicates the two distributions are identical.


\subsection{System-level Preference}
\label{sec:sys-pref}

The standard evaluation protocol in NLG consists of comparing the ranking of the systems based on the averaged evaluation scores. In many use cases, human and automatic metrics are in agreement about the system ranking. However, a prior study in Controlled Generation \cite{Dathathri2020Plug} shows that the assumption does not necessarily hold. Therefore, we design a task to compare the system ranking between automatic metrics and human as a similarity measure. 

\paragraph{Definition 1. System-level preference} Let $a$ and $b$ denote two independent NLG systems. We adopt the concept of utility function in human evaluation \cite{ethayarajh-jurafsky-2022-authenticity} to measure system-level preference. The relation $a \prec b$ means that $b$ is strictly preferred than $a$ if and only if the utility of $a <$ the utility of $b$:
\vspace{-.5em}
\begin{equation}
    \begin{aligned}
    a \prec b \iff u(a) < u(b).
    \end{aligned}
    \vspace{-.5em}
\end{equation}

\noindent $a \succ b$ means that $a$ is preferred than $b$, while $a \sim b$ means that $a$ and $b$ are indiscernible. In this study, the utility function $u(.)$ is the averaged evaluation scores for a particular NLG system.

\paragraph{Distance Measure}

To compute preference similarity between two metrics, we adopt Levenshtein distance, which calculates the minimum number of insertions, deletions, and substitutions required to change one sequence into the other sequence.

\begin{equation}
    \begin{aligned}
    d_i(\hat{P}, P) = \textnormal{Lev} (\hat{P}, P),
    \end{aligned}
\end{equation}

\noindent where $P$ and $\hat{P}$ can be expressed as two sequential orders of system-level preference. For example, let consider $P=a \prec b $ and $\hat{P} = b \prec a$. Then, Levenshtein distance between $\hat{P}$ and $P$ is $2$.

One of the limitations of Levenshtein distance is that the metric mainly calculates number of operations and does not take into account the sequence length differences. For example, the distance between two pairs $P_1 = \{cdabe, abcde\}$ and $ P_2 = \{cbed , abcde\}$ are same, $4$, even though the two pairs are composed of different sequences. To tackle this issue, we extend the distance metric formulation into a similarity measure by incorporating the total length of both sequences.

\paragraph{Definition 2. Preference similarity} The similarity $S$ between the two sequences $P_1$ and $P_2$ can be defined as:
\vspace{-.5em}
\begin{equation}
    \begin{aligned}
    S = \frac{(L_1 + L_2) - 2*\textnormal{Lev}(P_1, P_2)}{(L_1 + L_2)},
    \end{aligned}
\end{equation}

\noindent where $S$ denotes the similarity score; $L_1$ and $L_2$ are the length of $P_1$ and $P_2$ respectively. Using the above formula, the similarity between the first example pair $P_1 = \{cdabe, abcde\}$ is $0.2$, while the similarity of the second pair $ P_2 = \{cbed , abcde\}$ is $0.11$.

\subsection{Aspect-level Evaluation}

NLG evaluation involves addressing qualitative questions, such as ``Can the automatic metrics identify aspect-specific quality that is inferred in the generated texts?'' For example, a dialogue system that uses the preceding conversation as a context when generating a new question or response can be considered more engaging and more faithful to the context than the system that outputs repetitive responses. Thus, an automatic metric can be considered adequately \emph{good} if it can discern between low and high-quality system outputs. For measuring the capability of metrics on discerning aspect-level qualities, we utilize \textbf{Kolmogorov-Smirnov (KS)}, as described in Eq.~\ref{eq:ks}.

\begin{figure*}[!ht]
    \centering
    \begin{subfigure}[!t]{.45\textwidth}
         \centering
         \includegraphics[width=\linewidth]{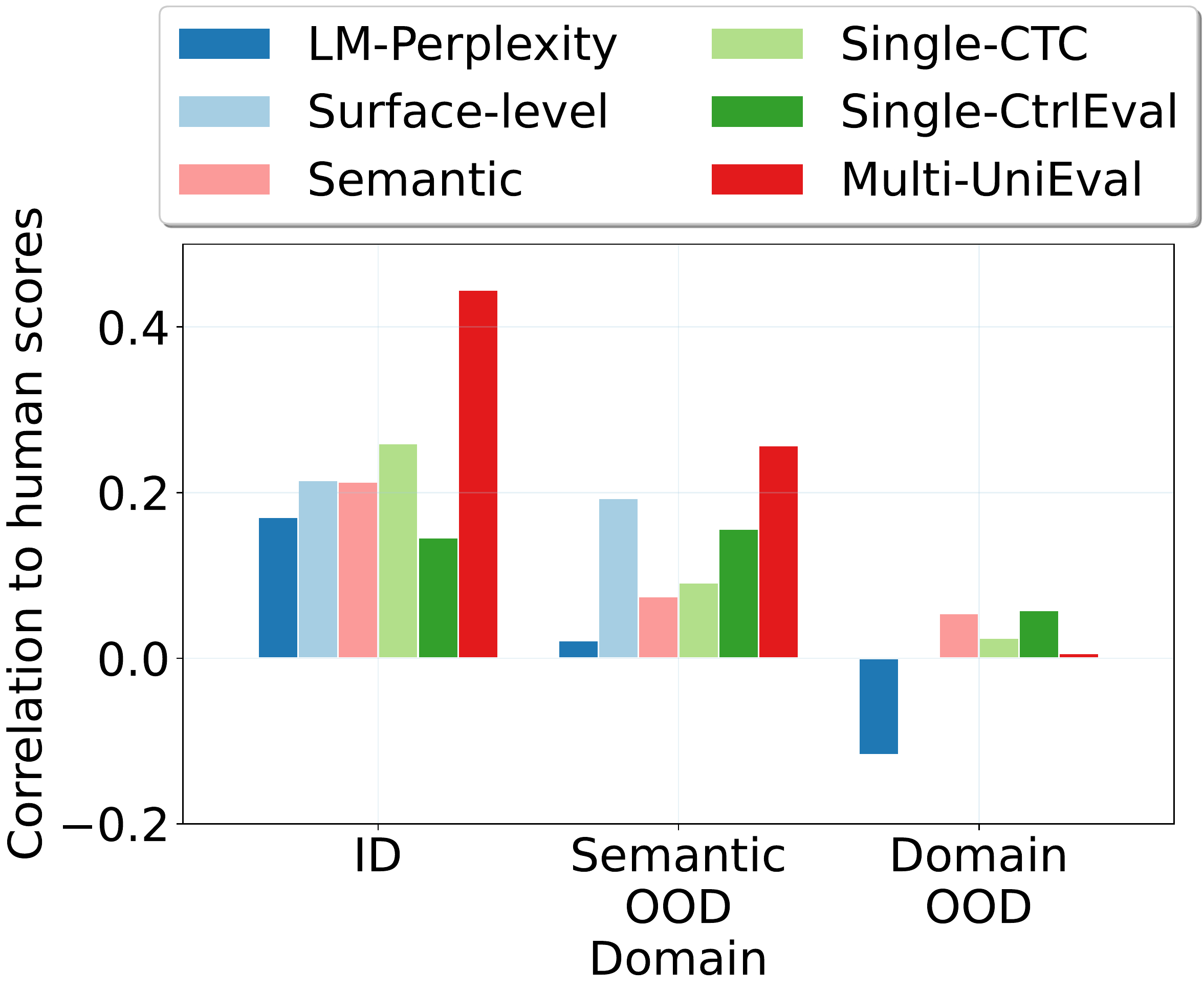}
         \caption{Correlation across ID and OOD.}
	\label{fig:transfer-ood}
       \end{subfigure}
    \begin{subfigure}[!t]{.45\textwidth}
         \centering
         \includegraphics[width=\linewidth]{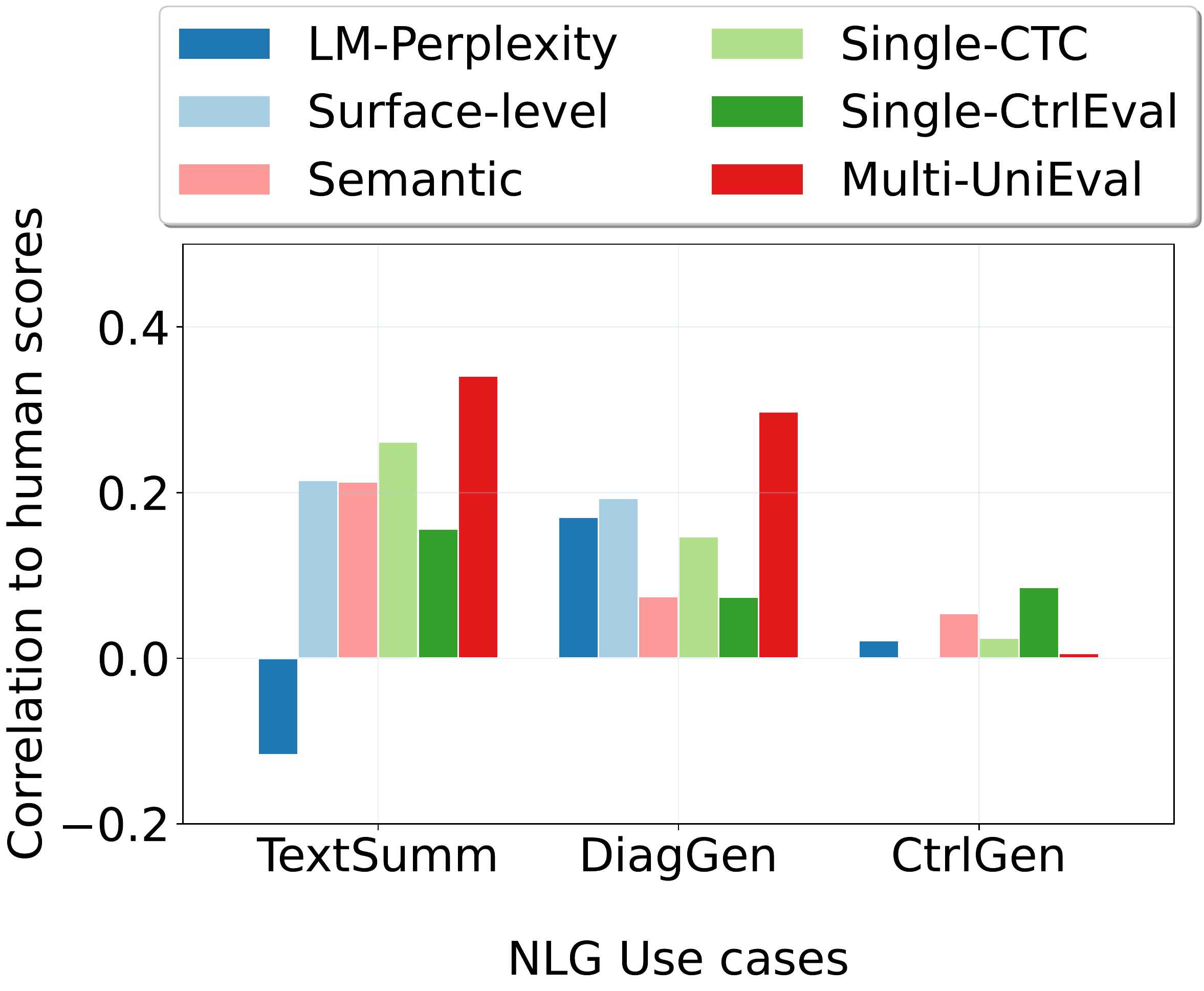}
         \caption{Correlation across NLG tasks. }
	\label{fig:transfer-task}
       \end{subfigure}
    \caption{Correlation level to human in transfer experiments (Zero-shot).}
    \label{fig:corr-transfer}
\end{figure*}

\section{Experiment}

\subsection{Datasets and Metrics \footnote{Details are provided in Appendix.}}

We consider publicly available author-annotated benchmark datasets in three NLG tasks, as listed in Table~\ref{tab:humeval-data}. For automatic metrics, we consider commonly used task-agnostic automatic metrics in NLG evaluation and the recent proposal of human-aligned automatic metrics, as listed in Table ~\ref{tab:stat-met}.

\subsection{Evaluation Setup}

\paragraph{ID vs OOD samples}
We classify benchmark datasets as target evaluation data into In-Domain (ID) and Out-of-Domain (OOD) categories, as shown in Table ~\ref{tab:stat-met}. The configuration of the data split is explained in section~\textsection~\ref{sec:transfer}.

\paragraph{Level of quality}
We split samples in each benchmark dataset into three categories (if applicable) based on their corresponding human ratings: \textbf{low} quality (rating $<3$); \textbf{moderate} (rating $=3$); and \textbf{high} quality (rating $>3$). The split is disjointly applied to each human evaluation aspect. 


\paragraph{Easy vs. Hard samples}
We split samples in each benchmark dataset into two categories: \textbf{Easy} and \textbf{Hard}. First, The rank of systems is obtained by averaging their corresponding human scores. \textbf{Easy} pair is composed of two systems with the large performance difference (e.g. systems with the lowest vs. highest human scores), while \textbf{Hard} pair contains systems with a close performance score.

\section{Results, Analysis, and Discussion}
\label{sec:results}

\subsection{Transfer Experiment}

Figure~\ref{fig:corr-transfer} shows the correlation level between automatic metrics and human ratings across NLG domains (ID and OOD). The result is summarized below.

\paragraph{Low Correlation in transfer experiment.} We observe that the correlation level of automatic metrics deteriorates sharply on target datasets with Semantic-Shift OOD and Domain-Shift OOD, particularly for tunable metrics, such as LM-based Perplexity, BERTScore, and human-aligned metrics (CTC, CtrlEval, UniEval). In general, the notably low correlation is observed in Controlled Generation (CtrlGen) task. \textbf{UniEval}'s correlation scores to human are considered moderately high in TextSumm (\textbf{0.341}) and DiagGen (\textbf{0.298}), but the metric does not correlate well with human in CtrlGen (\textbf{0.006}). The result suggests the remaining challenges of adapting human-aligned automatic metrics to a new task or domain, regardless whether the target task has similar dimensions of desirable human-like qualities.

\subsection{Aspect-level Evaluation}
\label{sec:aspect-level-eval}

\begin{figure*}[!ht]
    \centering
       \begin{subfigure}[t]{.48\textwidth}
         \centering
         \includegraphics[width=\linewidth]{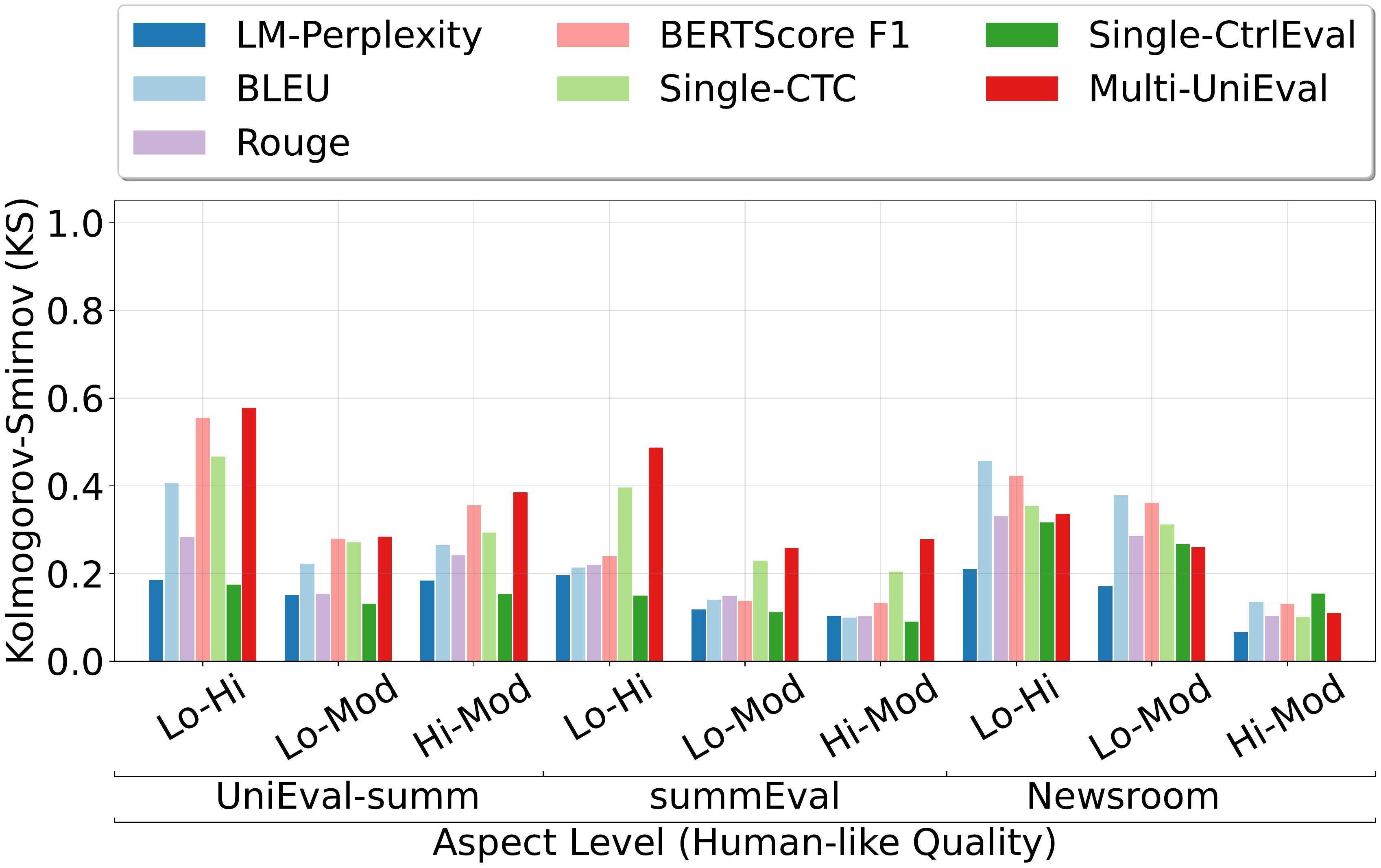}
         \caption{Identifying different levels of quality.}
         \label{}
       \end{subfigure}
    \hfill
    \begin{subfigure}[t]{.48\textwidth}
         \centering
         \includegraphics[width=\linewidth]{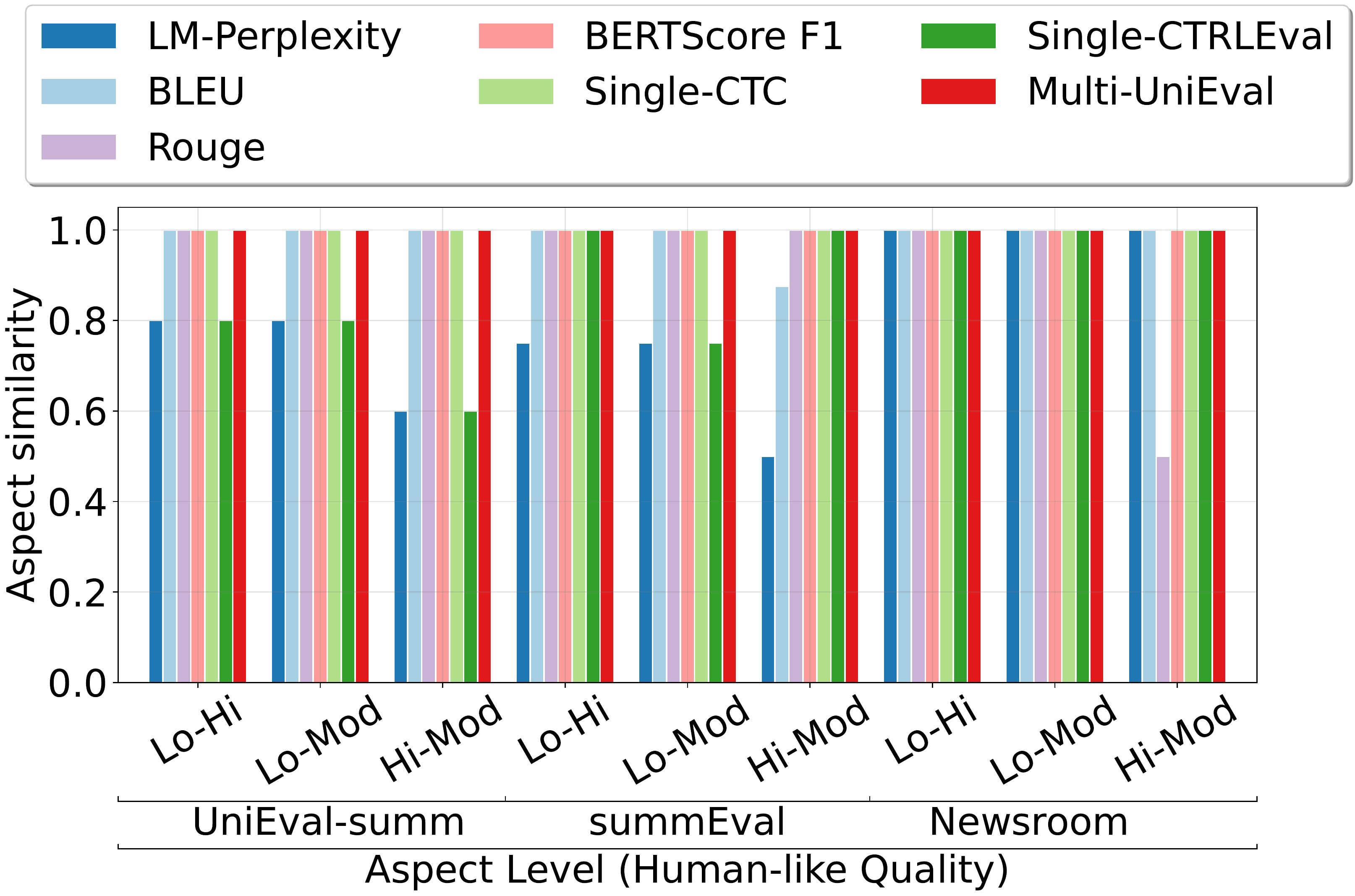}
         \caption{Rank/Preference similarity to human.}
         \label{}
       \end{subfigure}
    \caption{Aspect-level evaluation in Text Summarization (TextSumm). \textbf{Left}: Kolmogorov-Smirnov (KS) score on discerning between two different levels of human-like quality -- Higher is better $[0,1]$. \textbf{Right}: Similarity to the rank of the aspect-levels based on human scores -- Higher is better $[0,1]$. Lo-Hi: Low vs. High quality (e.g. Poor Coherent vs. Highly coherent), Lo-Mod: Low vs. Moderate. Hi-Mod: High vs. Moderate.
    }
    \label{fig:ks-summ-aspect}
\end{figure*}

\begin{figure*}[!ht]
    \centering
       \begin{subfigure}[t]{.48\textwidth}
         \centering
         \includegraphics[width=\linewidth]{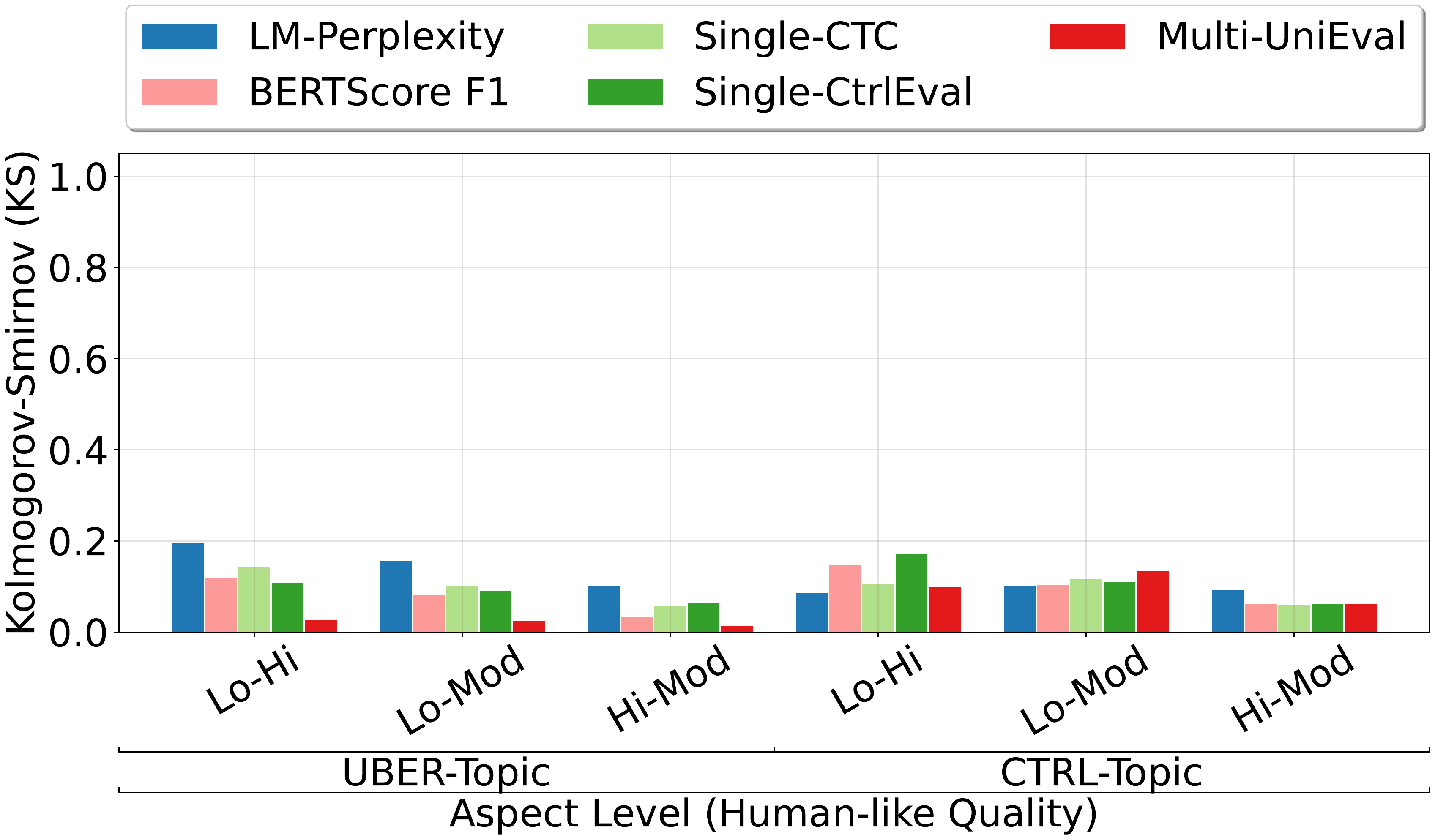}
         \caption{Identifying different levels of quality.}
         \label{fig:ks-ctrlgen-aspect1}
       \end{subfigure}
    \hfill
    \begin{subfigure}[t]{.48\textwidth}
         \centering
         \includegraphics[width=\linewidth]{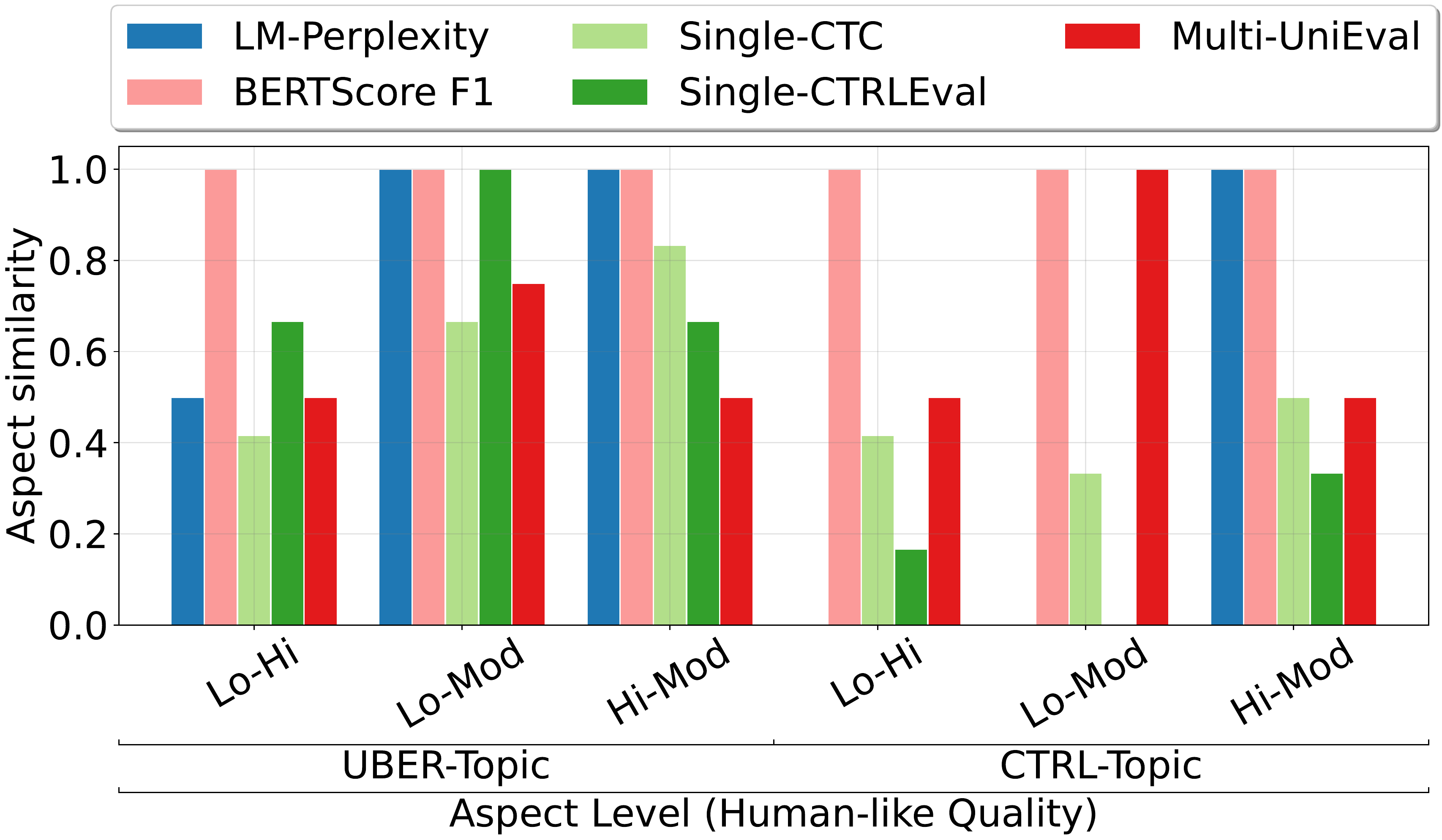}
         \caption{Rank/Preference similarity to human.}
         \label{fig:ks-ctrlgen-aspect2}
       \end{subfigure}
    \caption{Aspect-level evaluation in Controlled Generation (CtrlGen).
    }
    \label{fig:ks-ctrlgen-aspect}
    \vspace{-.5em}
\end{figure*}

Figure~\ref{fig:ks-summ-aspect}-\ref{fig:ks-ctrlgen-aspect} shows aspect-level evaluation of automatic metrics in Text Summarization (TextSumm) and Controlled Generation (CtrlGen). Our main observations are as follows:

\paragraph{UniEval performs best in TextSumm} Multi-aspect human-aligned metric (\textbf{UniEval}) is observed to have superior performance (up to \textbf{0.579}) at distinguishing between different levels of quality in UniEval-summ and summ-Eval. However, the discriminative power of the metric is less visible in Newsroom and Controlled Generation (CtrlGen) task. In Newsroom, both \textbf{BLEU} and \textbf{BERTScore} are more discriminative than human-aligned metrics (CTC, CTRlEval, UniEval).

\paragraph{BERTScore is comparably good in TextSumm} \textbf{BERTScore} has an adequately good discriminative property (KS=\textbf{0.557}) in UniEval-summ, comparable to multi-aspect human-aligned metric (\textbf{UniEval}) with KS=\textbf{0.579}. In Newsroom, \textbf{BERTScore} consistently has a higher performance score in three sample categories (Lo-Hi, Lo-Mod, Hi-Mod) than human-aligned metrics (CTC, CtrlEval, UniEval). The finding suggests that the characteristics of datasets in Text Summarization domain adequately fit with automatic metrics based on semantic similarity of text embeddings.

\paragraph{Higher KS is not necessarily highly agreeable}
\textbf{Perplexity} has the highest KS score for distinguishing between low and high quality outputs in UBER data. In contrast, the metric's aspect-level preference is not in alignment with human.

\subsection{System-level Evaluation}

Figure~\ref{fig:corr-transfer-summ}-\ref{fig:corr-transfer-ctrl} show the effectiveness of the metrics at discerning system-level performance. Our main observations are as follows:

\paragraph{BLEU is more discriminative in Newsroom}
In general, apart from \textbf{BLEU} in \textbf{Newsroom}, the remaining metrics' KS scores across three NLG tasks are considered low-to-moderate ($\leq 0.6$). We further inspect the reason why \textbf{BLEU} performs considerably well in Newsroom and discover that the data is mainly composed of outputs from two types of NLG systems: extractive vs. abstractive summarization systems. We also observe that in the Newsroom dataset, abstractive systems are often voted lower (averaged score = \textbf{2.5}) than extractive systems (averaged score =\textbf{3.85}). Such characteristic of human ratings in Newsroom is a good fit for surface-level metric (BLEU), because the metric is more likely to penalize abstractive systems with zero score (\textbf{0.0}) and extractive systems with a higher score (e.g. \textbf{1.0}).

\paragraph{Automatic metrics are more discriminating than human } When human struggles to distinguish between different system-level performances, automatic metrics are observed to be more discriminative. For example, in UniEval-summ (Hard), human has a considerably low score (KS =\textbf{0.145}), while \textbf{UniEval} has a higher KS score (KS =\textbf{0.269}). In Newsroom (\emph{Hard}), \textbf{BLEU}, \textbf{BERTScore}, and \textbf{UniEval} are more discriminative (KS $>$ \textbf{0.4}) than human (KS=\textbf{0.163}). The possible reason for this particular use case is that \emph{Hard} sample pairs are mainly composed of systems from a similar source or origin. For example, in Persona-Chat (USR-PC), the \emph{Hard} sample category is composed of a pair of human reference systems: \textbf{Original Ground Truth}, \textbf{New Human Generated}. In Newsroom, \emph{Hard} sample pairs consist of models from the same category (e.g. extractive-based systems). In UBER-Topic, where low KS scores are more visible across human and automatic metrics, both \emph{Easy} and \emph{Hard} pairs consist of systems that are derived from one pretrained Language Model.

\paragraph{Multi-aspect human-aligned metric is not always dominant}
In Persona-Chat (USR-PC), a single aspect human-aligned metric (\textbf{CTC}) has a higher KS score (\textbf{0.386}) and higher preference similarity (\textbf{0.888}) than a multi-aspect metric (\textbf{UniEval}), in which KS =\textbf{0.218} and similarity=\textbf{0.833}. In UBER-Topic, UniEval has the lowest KS score (\textbf{0.025} for Easy pairs, \textbf{0.027} for Hard pairs). We find that the less distinctiveness of \textbf{UniEval} is mainly due to a high alignment between \textbf{UniEval} and multi-dimensional human evaluation aspects. For example, in Persona-Chat (USR-PC), the agreement between human evaluation aspects is low. The three aspects (\emph{Understandable, Natural, and Engaging}) yield a different system rank than the remaining aspects. Thus, a high alignment to inter-aspect disagreement may necessarily introduce a lower KS.




\subsection{Visualizing Pairwise System Ranking}
\label{sec:vis-pairwise}

\begin{figure*}[!ht]
    \centering
       \begin{subfigure}[t]{.48\textwidth}
         \centering
         \includegraphics[width=\linewidth]{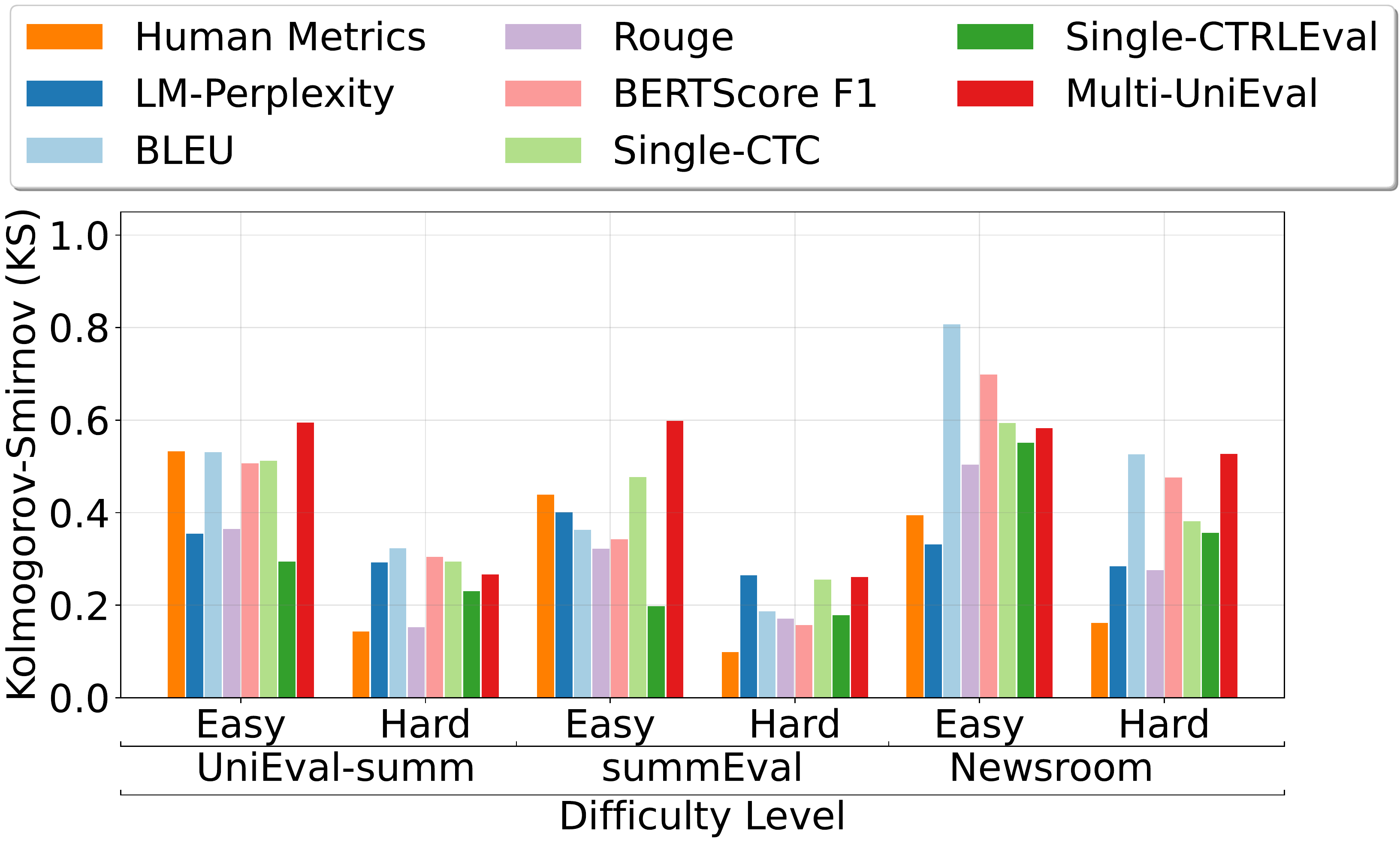}
         \caption{Identifying system-level performance difference.}
         \label{}
       \end{subfigure}
    \hfill
    \begin{subfigure}[t]{.48\textwidth}
         \centering
         \includegraphics[width=\linewidth]{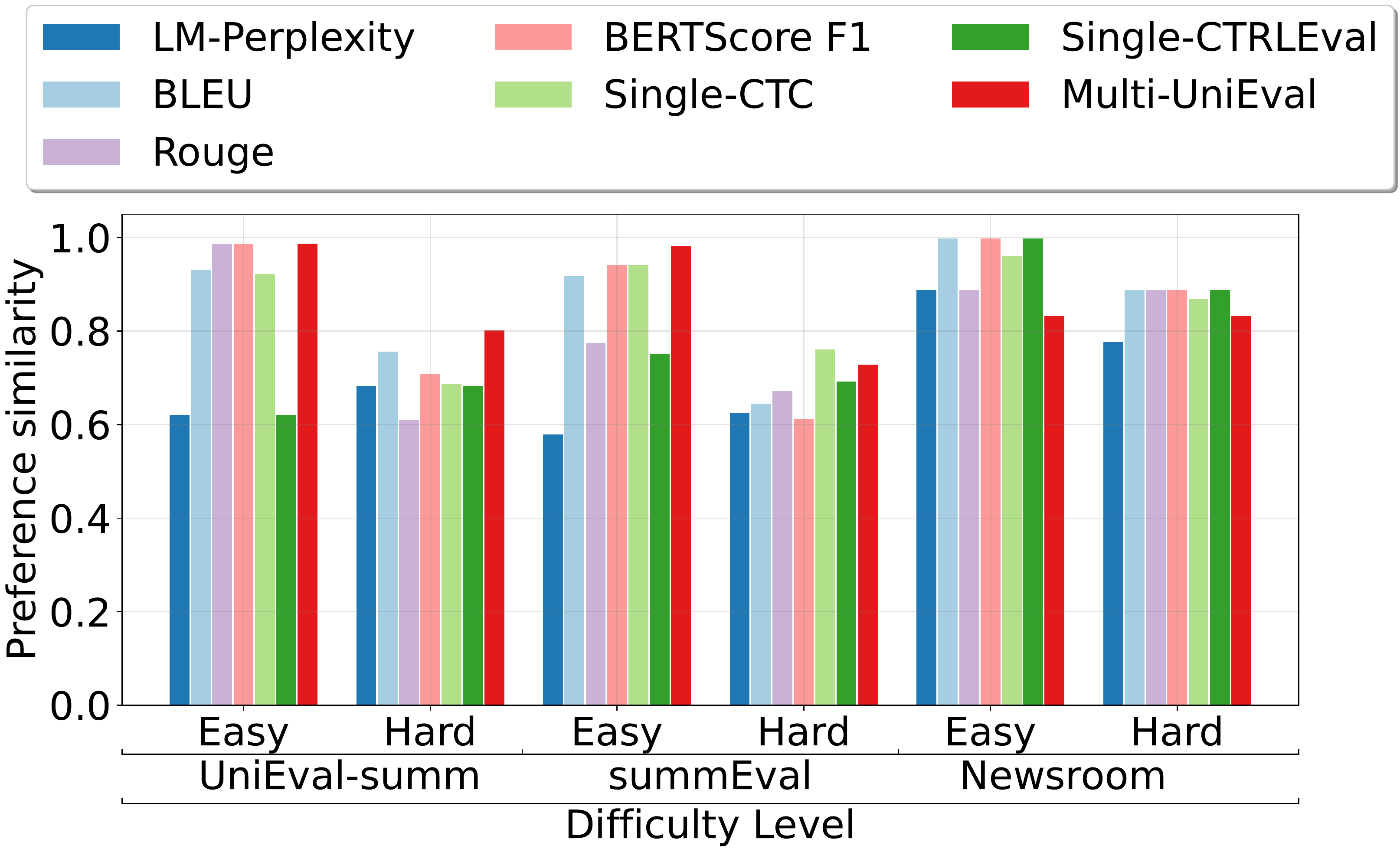}
         \caption{Rank/Preference similarity to human.}
         \label{}
       \end{subfigure}
    \caption{System-level evaluation in Text Summarization (TextSumm). \textbf{Left}: Kolmogorov-Smirnov (KS) score on discerning the performance difference between two independent NLG systems -- Higher is better $[0,1]$. \textbf{Right}: Preference similarity between human and automatic metric -- Higher is better $[0,1]$. 
    }
    \label{fig:corr-transfer-summ}
\end{figure*}
\begin{figure*}[!ht]
    \centering
       \begin{subfigure}[t]{.48\textwidth}
         \centering
         \includegraphics[width=\linewidth]{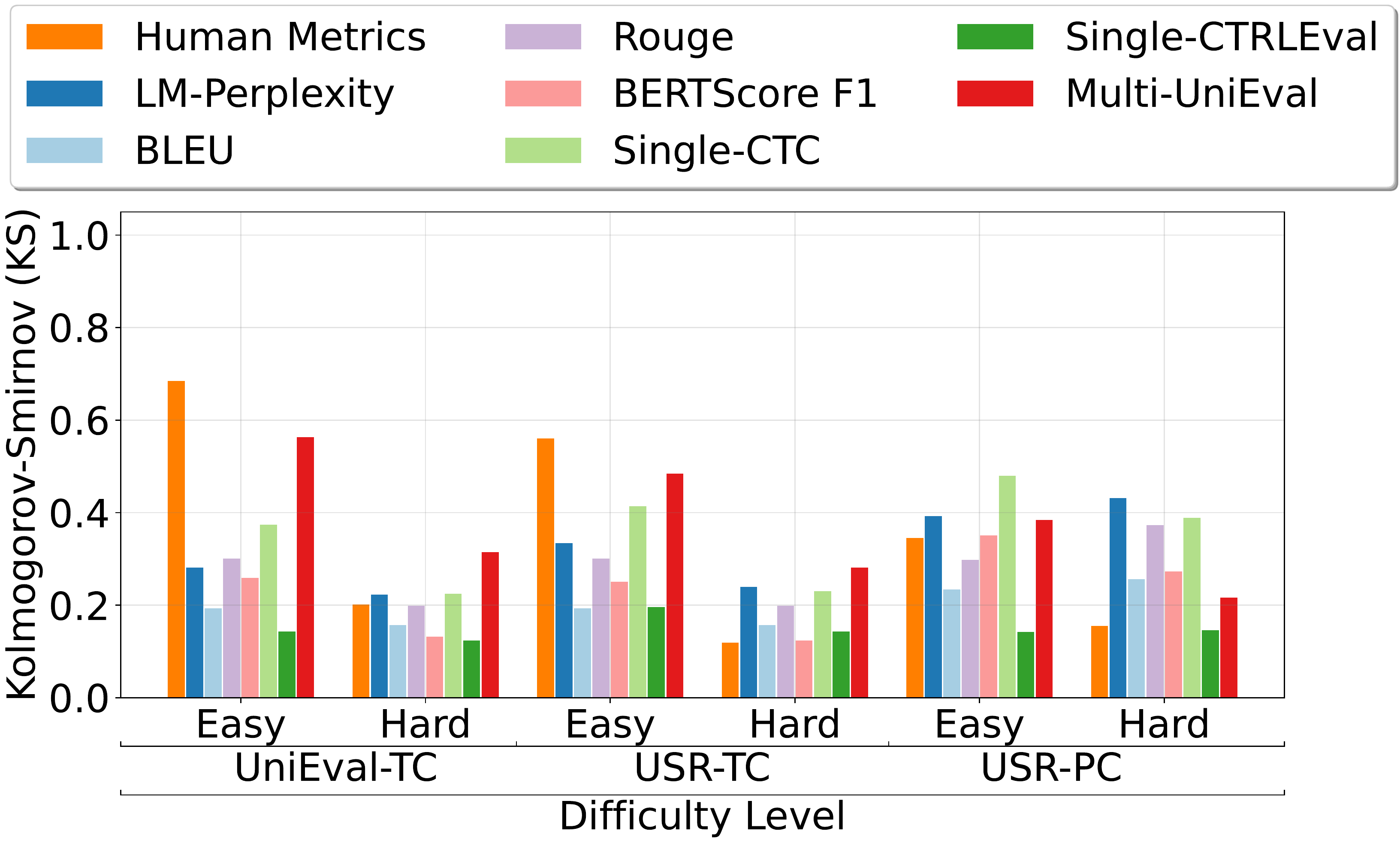}
          \caption{Identifying system-level performance difference.}
         \label{}
       \end{subfigure}
    \hfill
    \begin{subfigure}[t]{.48\textwidth}
         \centering
         \includegraphics[width=\linewidth]{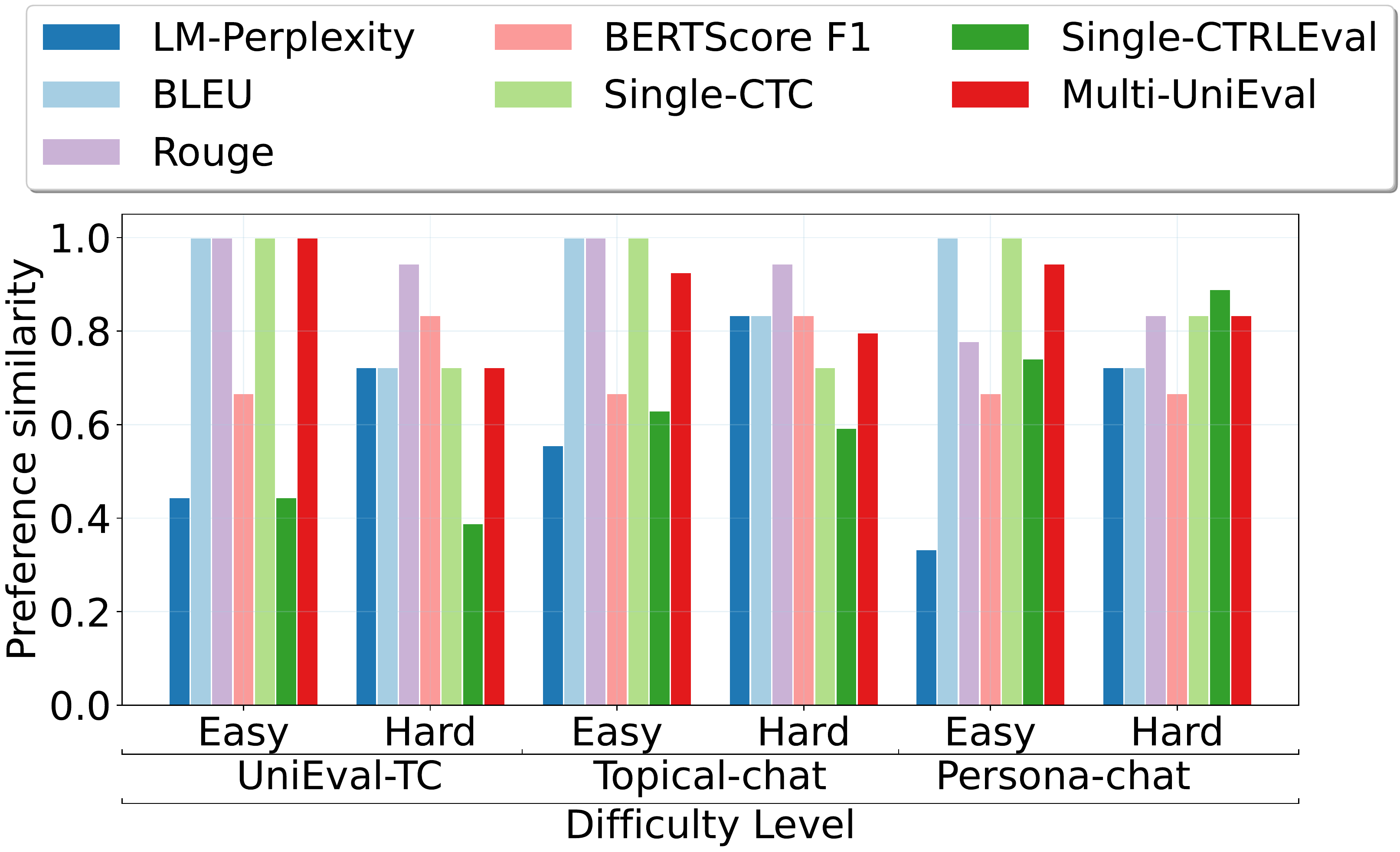}
         \caption{Rank/Preference similarity to human.}
         \label{}
       \end{subfigure}
    \caption{System-level evaluation in Dialogue Response Generation (DiagGen).}
    \label{fig:corr-transfer-diag}
\end{figure*}
\begin{figure*}[!ht]
    \centering
       \begin{subfigure}[t]{.48\textwidth}
         \centering
         \includegraphics[width=\linewidth]{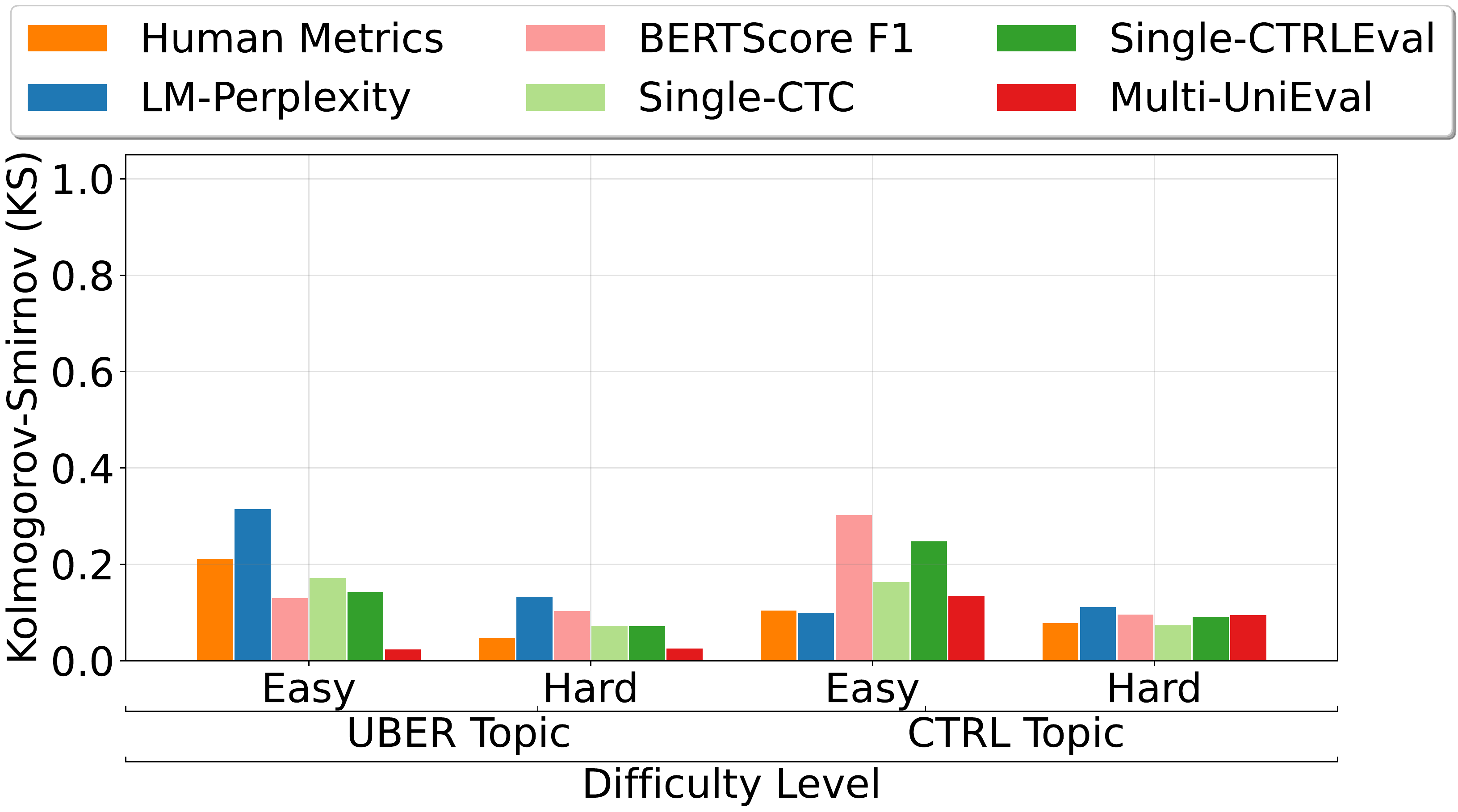}
          \caption{Identifying system-level performance difference.}
         \label{}
       \end{subfigure}
    \hfill
    \begin{subfigure}[t]{.48\textwidth}
         \centering
         \includegraphics[width=\linewidth]{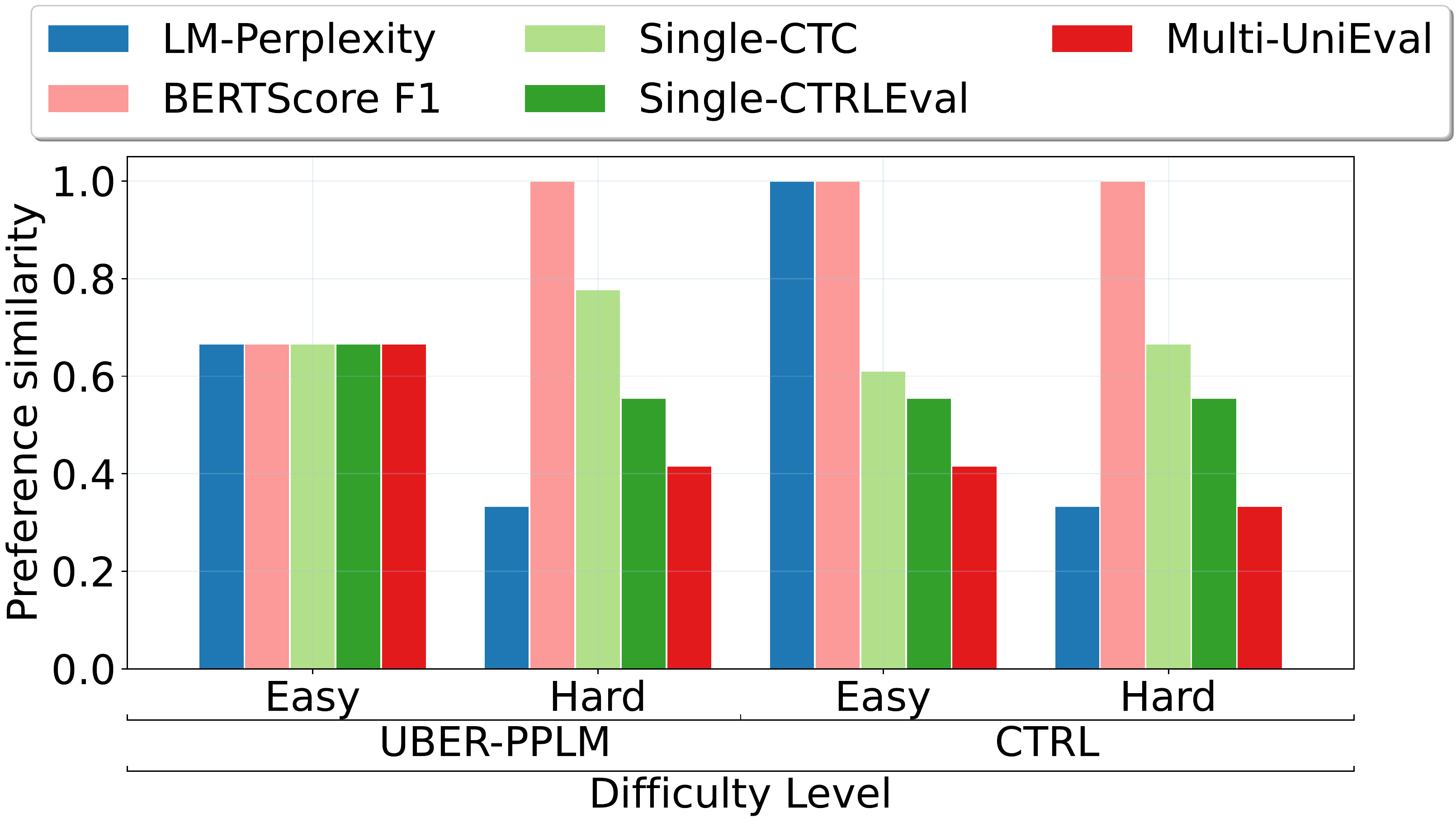}
         \caption{Rank/Preference similarity to human.}
         \label{}
       \end{subfigure}
    \caption{System-level evaluation in Controlled Generation (CtrlGen). }
    \label{fig:corr-transfer-ctrl}
\end{figure*}

We compare pairwise win fractions of NLG systems based on human ratings and automatic metrics in this study. The objectives are: (i) to better reason on why automatic metrics are more discriminating than human and (ii) to inspect the agreement level between metrics on system ranking.

\begin{figure*}[!ht]
    \centering
    \begin{subfigure}[t]{.21\textwidth}
         \centering
         \includegraphics[width=\linewidth]{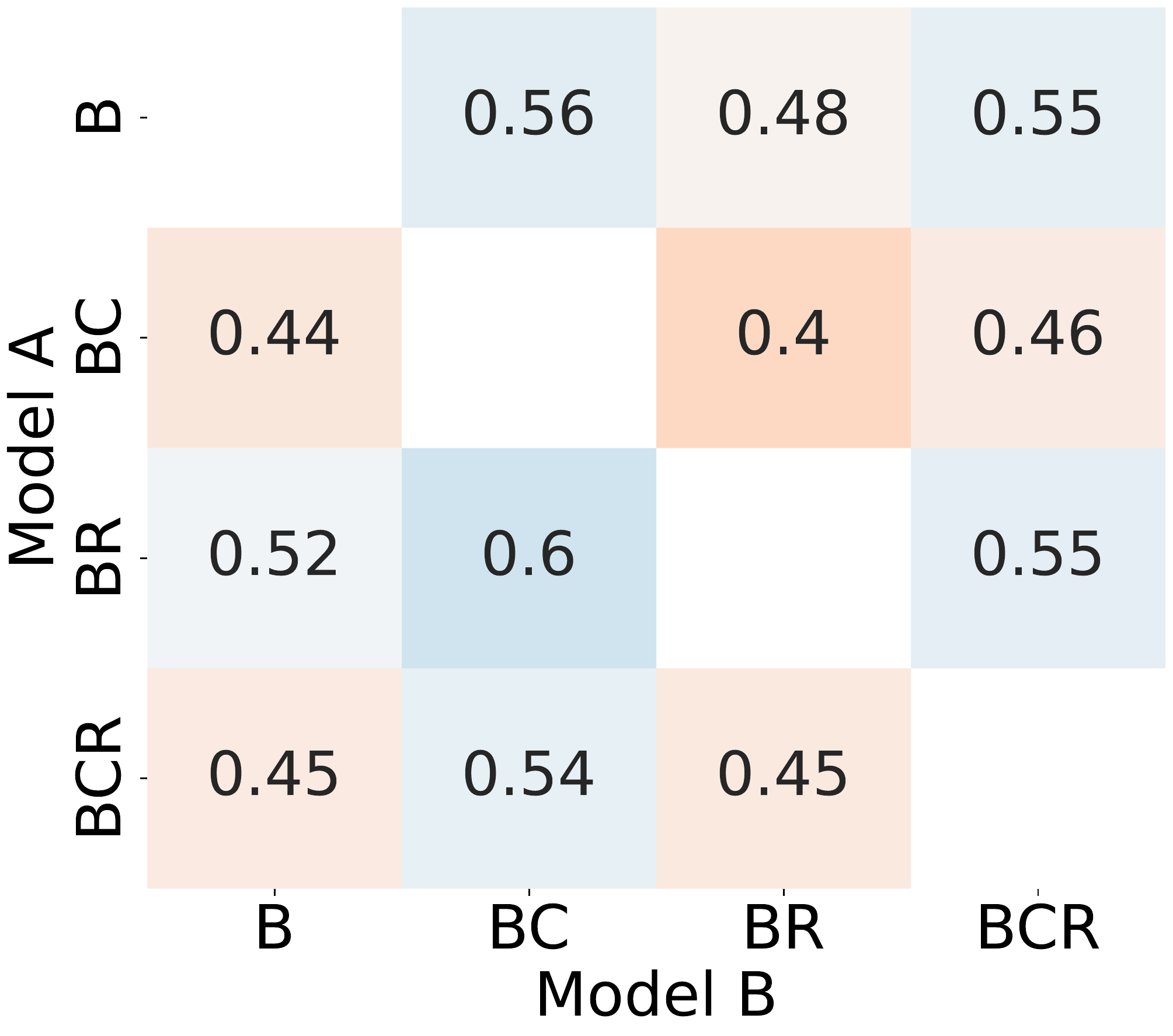}
         \caption{Fluency (Human)}
	   \label{fig:fluency-win}
    \end{subfigure}
    \hfill
    \begin{subfigure}[t]{.21\textwidth}
         \centering
         \includegraphics[width=\linewidth]{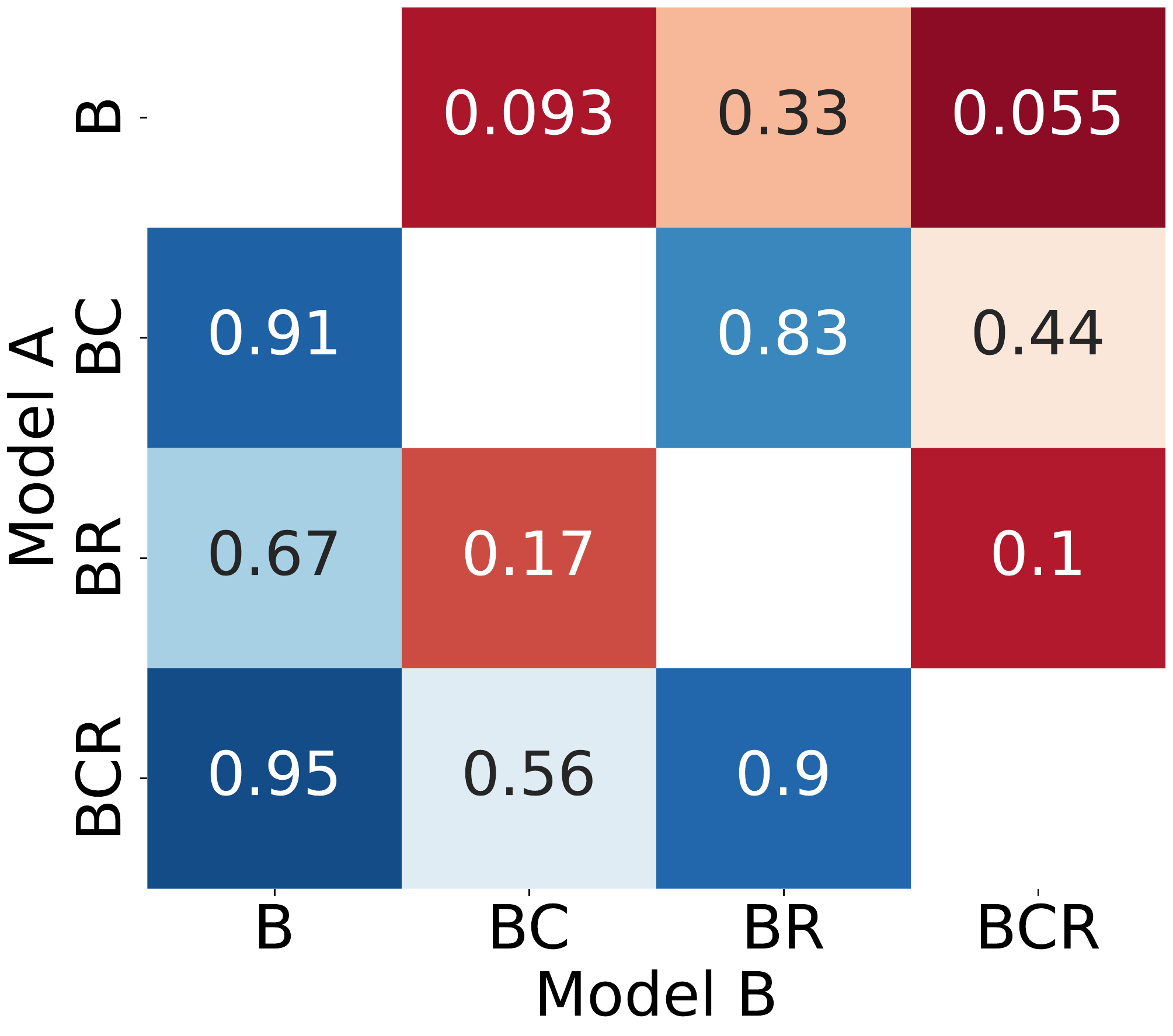}
         \caption{Relevance (Human))}
	   \label{fig:relev-win}
    \end{subfigure}
    \hfill
    \begin{subfigure}[t]{.21\textwidth}
         \centering
         \includegraphics[width=\linewidth]{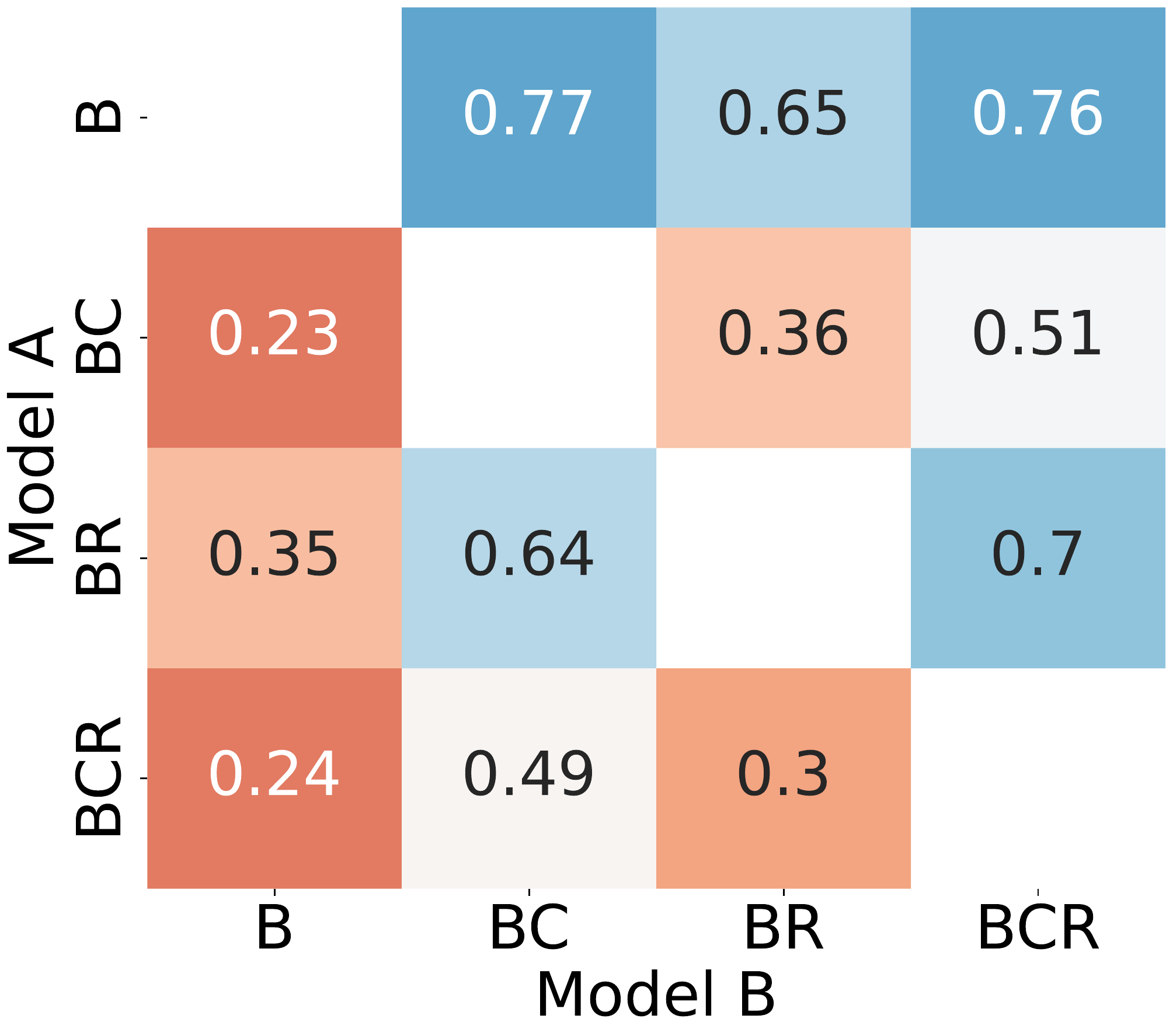}
         \caption{Perplexity}
	   \label{fig:perplex-win}
    \end{subfigure}
    \hfill
    \begin{subfigure}[t]{.21\textwidth}
         \centering
         \includegraphics[width=\linewidth]{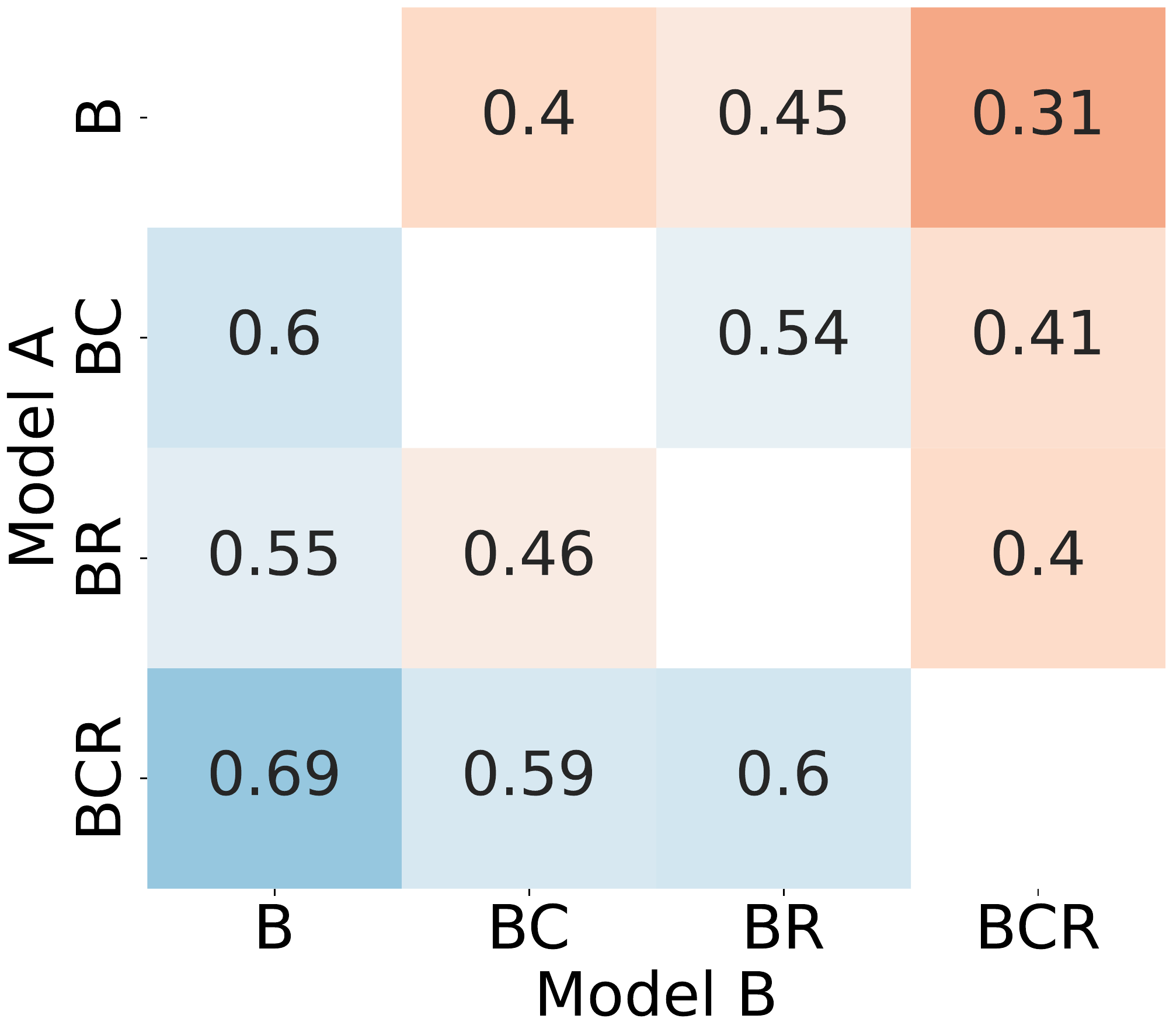}
         \caption{BERTScore F1}
	   \label{fig:bertsc-win}
    \end{subfigure}
    \hfill
    \begin{subfigure}[t]{.21\textwidth}
         \centering
         \includegraphics[width=\linewidth]{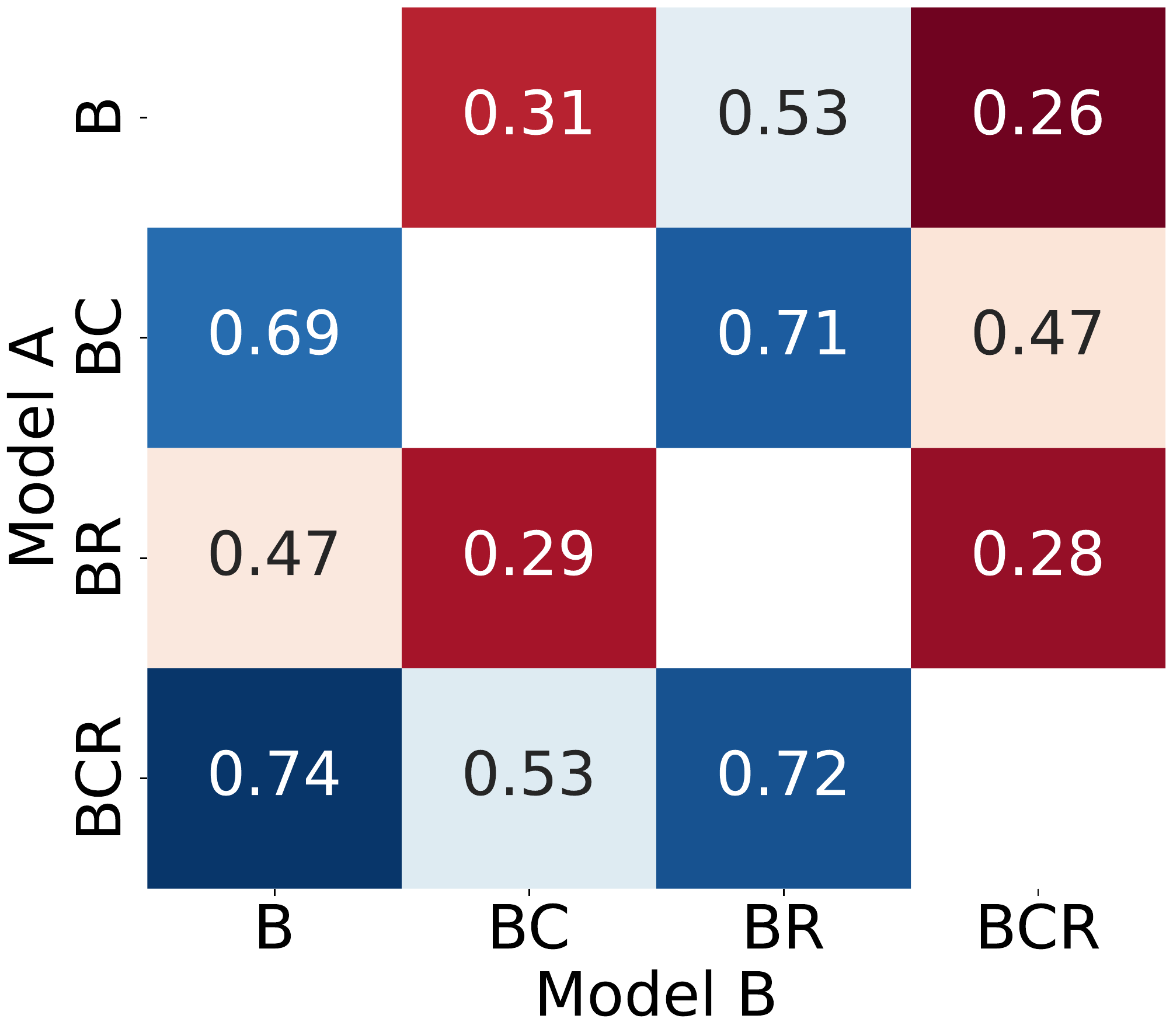}
         \caption{CTC-E Relevance}
	   \label{fig:ctce-relev-win}
    \end{subfigure}
    \hfill
    \begin{subfigure}[t]{.21\textwidth}
         \centering
         \includegraphics[width=\linewidth]{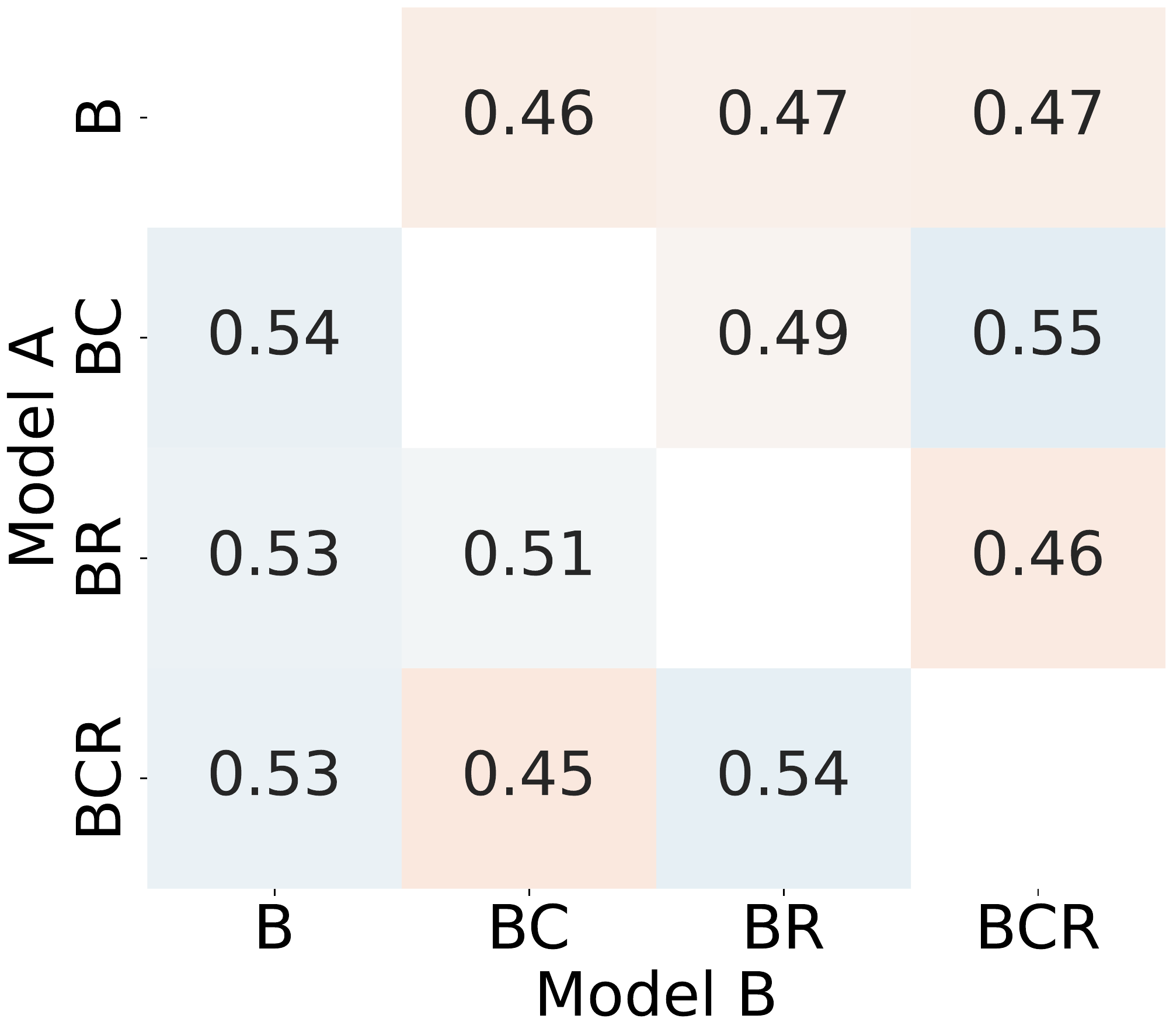}
         \caption{CTRLEval Relevance}
	   \label{fig:ctrleval-relev-win}
    \end{subfigure}
    \hfill
    \begin{subfigure}[t]{.21\textwidth}
         \centering
         \includegraphics[width=\linewidth]{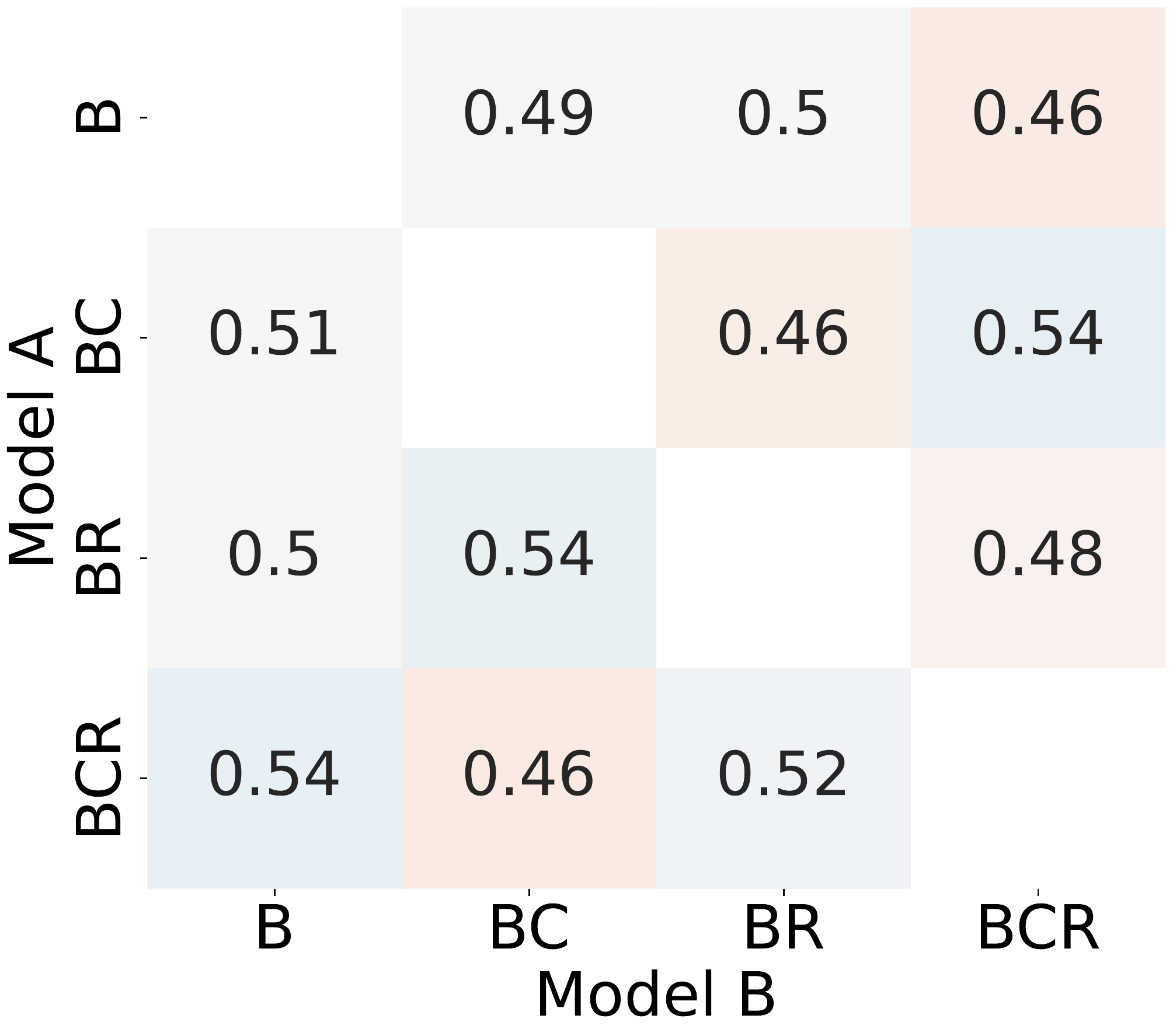}
         \caption{UniEval-Fluency}
	   \label{fig:unieval-overal-win}
    \end{subfigure}
    \hfill
    \begin{subfigure}[t]{.21\textwidth}
         \centering
         \includegraphics[width=\linewidth]{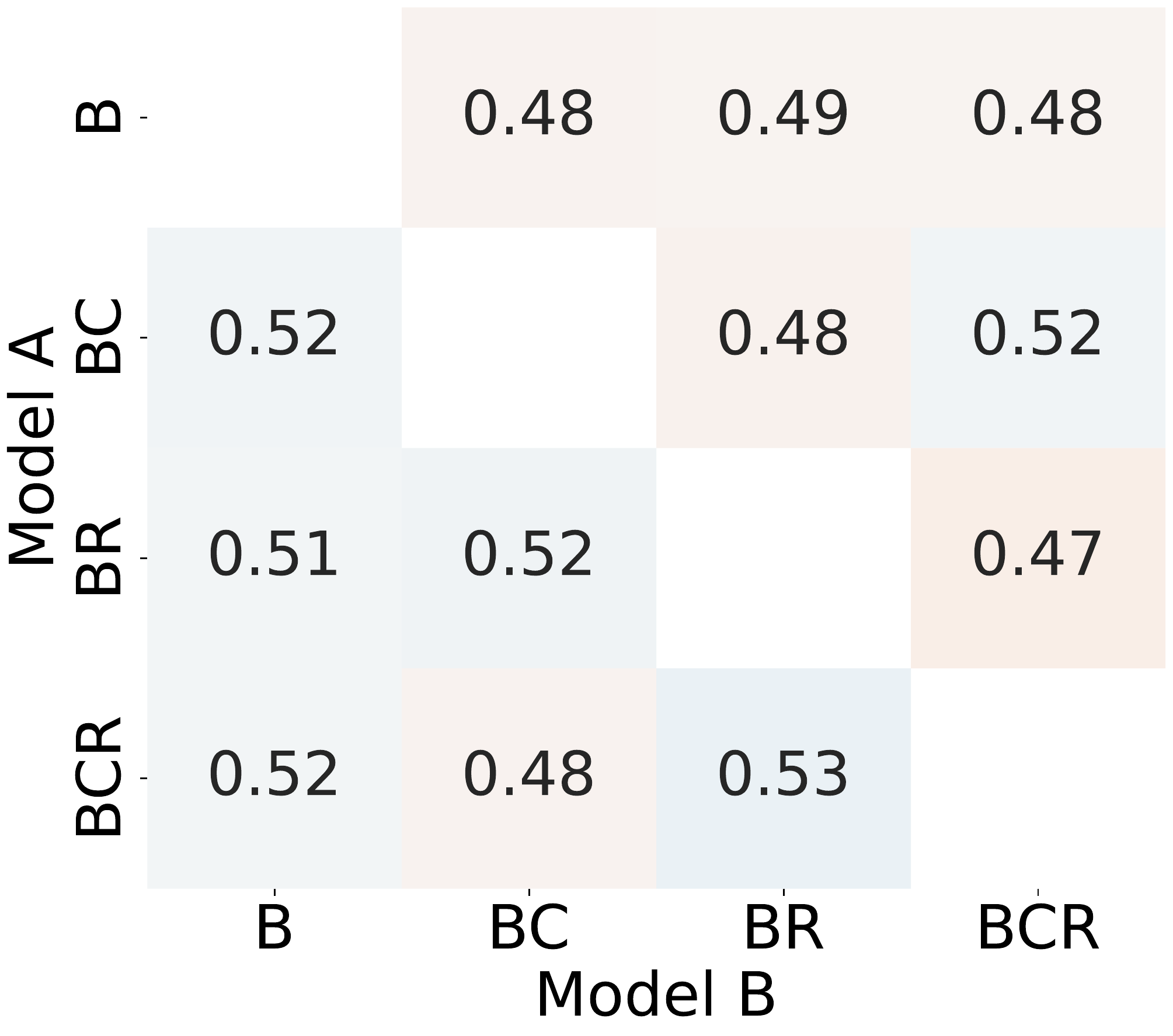}
         \caption{UniEval-Relevance}
	   \label{fig:unieval-relev-win}
    \end{subfigure}
    \caption{Pairwise win fractions in Controlled Generation (UBER PPLM data, Control: Topic). The number represents fraction of Model A wins over model B. Order matters here because human evaluators are asked to rate systems based on random pairings: System A-B can be represented as both A-B and B-A. $\uparrow$ \textbf{Higher} is \textbf{better}.}
    \label{fig:pairwise-wins}
\end{figure*} 

Notice that the results of pairing evaluation, as shown in Figure~\ref{fig:pairwise-wins}, are consistent with our empirical findings in Figure~\ref{fig:corr-transfer-summ}-\ref{fig:corr-transfer-ctrl}, particularly for preference similarity with human. The system rankings based on BERTScore F1 and single-aspect CTC metrics are more similar to human on \emph{Relevance}. Perplexity is more discriminating than human, but its similarity to human (\emph{Fluency}) is lower. We also observe that although automatic metrics are more discriminating than human ratings in general, human voting on \emph{Relevance} aspect can discern system-level performance more effectively than BERTScore and CTC-E Relevance. The result suggests that although a binary voting scheme in a human evaluation study may be less insightful than rating or error correcting protocol, the approach is cost-effective for performance selection based on a particular evaluation aspect.


\section{Implications}
\label{sec:implications}


\subsection{Faithfulness to Human Preference}
\label{sec:faithful}

We show that both low correlation scores and low discriminative power (KS scores) do not represent low faithfulness to human preference. In Controlled Generation, we observe that metrics with lower correlation and lower KS score, such as BERTScore-F1 and single-aspect CTC, on the contrary have a higher similarity with human on system-level preference and ranking. The result suggests the importance of verifying the metric's correlation score to its faithfulness to human preference, particularly for NLG use cases with poor correlation score (e.g. $\rho < 0.2$) and low agreement on system ranking.

\subsection{Discriminating System-level Performance}
\label{sec:discriminate}

We show that automatic metrics can be more discriminating than human, particularly when NLG systems are derived from the same training objective or encoding scheme. In contrast, for human evaluation aspect that is measured based on a binary voting scheme, such as \emph{Relevance} in Controlled Generation, we observe that the scores based on the corresponding aspect are more distinctive than automatic metrics. 



\subsection{Guidance to System Selection}
\label{sec:benchmark}

We show that benchmarking NLG systems and evaluation metrics via pairwise comparison provides more insights into the agreement level for selecting the best-performed system. Low agreement between metrics on ranking system-level performance suggests at least two scenarios. \textbf{First}, the automatic metrics are not able to capture the human-like qualities inferred in texts as key factors for discriminating system outputs. \textbf{Second}, each metric focuses on a particular evaluation aspect among multi-dimensional human-like qualities. For example, \emph{Fluency} focuses on penalizing repetition and grammatical errors, while \emph{Relevance} focuses on measuring the closeness between the generation outputs and the given control attribute (e.g. topic category). For guiding the selection of the best-performed system, the second scenario allows a fine-grained assessment to scrutinize both strengths and limitations of the system based on desirable human-like qualities.  

\section{Conclusion}

We introduce the metric preference checklist as a framework for analyzing the effectiveness of currently available NLG automatic metrics. We show the importance of verifying the preference similarity between automatic metrics and human, regardless of their correlation scores. We also find that automatic metrics are more discriminating than human for discerning system-level performance, except for human evaluation aspect with a binary voting protocol. Lastly, we show the implication of current work on guiding the selection of the best-performed system based on pairwise system ranking.

\section*{Limitations}

\paragraph{Robustness to perturbations} Our empirical study does not explore the connection between the discriminative power of automatic metrics based on the proposed metric preference checklist and their robustness to simple perturbations or other natural language phenomena that may occur in texts or NLG use cases.

\paragraph{Metric Fairness (Social Bias)} Our study does not include an investigation of metric fairness or social bias issues that may be introduced by Language Model-based NLG evaluation Metrics. 

\paragraph{Single-aspect vs. Multi-aspect} Our current empirical experiments mainly explore the discriminative power of evaluation metrics in single-aspect experiment setup (section \textsection{\ref{sec:aspect-level-eval}). It may also be interesting to inspect to what extend the metrics can identify multi-aspect levels of quality, particularly when there exists disagreement between human evaluation aspects. For example, instead of disjointly splitting samples into \{low \emph{Engagingness}, moderate \emph{Engagingness}, high \emph{Coherence}\}, samples can be divided based on the joint aspects, such as \{low \emph{Engagingness} and low \emph{Coherence}\}.

\paragraph{Universal input-output structure} Our experiments are mainly carried on publicly available author-annotated human evaluation benchmark datasets. Thus, we do not guarantee the universal input-output structure and a uniform naming system across datasets or tasks. For example, UniEval - Topical Chat data (UniEval-TC) \cite{zhong-etal-2022-towards} and USR - Topical Chat (USR-TC) \cite{mehri-eskenazi-2020-usr} use a different naming system for human evaluation aspects, yet the aspects refer to the same dimension of human-like qualities.

\paragraph{Dependency of NLG Systems} When comparing outputs from two different NLG systems, the systems are presumably independent. However, in many NLG use cases, this assumption is not fully accurate. For example, in Controlled Generation task, the systems originate from one pretrained Language Model as an encoder model. In inference or decoding stage, the encoder's probability outputs are used as inputs for multiple decoding schemes, such as the use of Log-Likelihood ranking, distance scoring as filter, etc \cite{Dathathri2020Plug}, yielding $n$-systems to compare with. As a result of this setup, the generation outputs from these $n$-systems are often less diverse and less distinguishable than the outputs from two independent systems that do not share the same encoding scheme or training objective.

\section*{Ethics Statement}
The purpose of this study is not to provide an immutable checklist to define what makes a good NLG evaluation metrics. Instead, the main objective is to introduce an extended perspective on how to assess metric-level performance beyond a correlation analysis. Our empirical experiments are carried on previously reported human evaluation data and NLG use cases under ACL Ethics Policy. Human evaluation datasets are extracted from peer-reviewed scientific publications by \citet{mehri-eskenazi-2020-usr} in ACL 2020; \citet{Dathathri2020Plug} in ICRL 2020; \citet{ke-etal-2022-ctrleval} in ACL 2022; and \citet{zhong-etal-2022-towards} in EMNLP 2022, as we have listed in our Experiment section.  Our empirical findings are not necessarily representative for NLG use cases and datasets that are not covered in this study. However, our metric preference checklist can be easily adopted as fine-grained analysis to measure the effectiveness of new NLG automatic evaluation metrics, regardless of their overall correlation scores to human judgments.

\section*{Acknowledgment}
We thank the anonymous reviewers for the constructive feedback, which has greatly improved the final version of the paper. This research has been partially supported by the Dutch Research Council (NWO) and Indonesian Endowment Fund for Education (LPDP) Scholarship under Beasiswa Pendidikan Indonesia (BPI) -- ID Number 0003194/SC/D/9/LPDP2016. The content of the information does not necessarily reflect the position or the policy of the Government, and no official endorsement should be inferred.

\bibliography{anthology,custom}
\bibliographystyle{acl_natbib}

\appendix

\section{Appendix}
\label{sec:appendix}

\subsection{Modification Post Reviews}

We thank reviewers for the constructive feedback. We list the modification of the paper based on reviewers' suggestion as follows.

\begin{itemize}
    \item We add the visualization of pairwise system ranking (section \textsection{\ref{sec:vis-pairwise}}) to accomodate the reviewers' suggestion on linking the current work to the objectives of NLG evaluation, particularly for reasoning and guiding model selection,
    \item We add \textbf{Implications} (\textsection{\ref{sec:implications}}) to improve the clarity of the paper,
    \item We add \textbf{Related Work} in the main page (section \textsection{\ref{sec:related-work}}) to clarify the positioning of current proposed framework, 
    \item We add \textbf{Background} in Appendix for providing detail information on NLG tasks and automatic metrics used in this study.
    \item We improve the presentation of the paper by highlighting the core points and the implications of the study for future works. We also correct the grammatical errors found in the manuscript. The revision is particularly done for \textbf{Abstract}, \textbf{Introduction}, \textbf{Related Work}, and \textbf{Conclusion} section.
    
\end{itemize}

\subsection{Background}

\subsubsection{NLG Tasks}
\label{sec:nlg-cases}

Our empirical study is mainly carried on three (3) NLG tasks: Controlled Generation, Dialogue Response Generation, and Text Summarization.

\paragraph{1. Controlled Generation} (CtrlGen) \cite{Dathathri2020Plug} is firstly introduced as Conditional Language Modeling \cite{Keskar2019CTRLAC}. In a general setup of CTRLGen, NLG systems are mainly trained based on a language modeling objective where the task is to predict next token or word given the preceding sequence of tokens. During inference stage, the trained system is perturbed with an external control attribute (e.g. topics, sentiment labels, aspects of sentiment) to generate texts that are semantically linked to the control attribute. All tasks in CtrlGen can be categorized as open-ended NLG tasks because ground truth human references are not provided by default. The quality of NLG system outputs is defined based on how semantically close the generation outputs to the corresponding control attribute, which can be aligned to several human-likeness aspects, such as \emph{coherence, consistency, fluency,} and \emph{relevance}. 

\paragraph{End-to-End NLG Systems} We measure the performance of the following systems based on previous work on in Controlled Generation task \cite{Dathathri2020Plug}: \textbf{B:} Baseline, unchanged pretrained GPT-2 Language Model. \textbf{BR:} Sampling B $r$ times based on Log Likelihood ranking and distance-based ranking. \textbf{BC:} For each decoding step, update latent representation $\tilde{H}_t$ based on attribute model log likelihood loss. \textbf{BCR:} Combine approach from BC (updating $\tilde{H}_t$) and BR (sampling and output ranking).

\paragraph{2. Dialogue Response Generation} (DiagGen) is NLG use case in neural conversational domain, which can be further divided into an investigation of multi-turn dialogue response generation in a Persona Chat domain \cite{see-etal-2019-makes}; or single response generation in Topical Chat and Persona Chat domains \cite{mehri-eskenazi-2020-usr,zhong-etal-2022-towards}. In this study, we focus on the evaluation of the latter category, where the quality of NLG system outputs is mainly assessed based on how good the machine responses to the preceding conversation. The \emph{goodness} is mainly defined based on several aspects of human-likeness, such as \emph{understandability, naturalness, coherence, engagingness,} and \emph{groundedness}.


\paragraph{End-to-End NLG Systems} For Persona-Chat dialogue response generation (USR-PC), we compare the performance of the following systems based on \cite{mehri-eskenazi-2020-usr,zhong-etal-2022-towards}: Systems based on pretrained models in ParlAI \footnote{\url{https://github.com/facebookresearch/ParlAI/tree/main/projects/convai2}} for CONVAI2 competition \cite{colombo2022infolm}, i.e. \textbf{Seq2Seq} -- a Sequence-to-Sequence model trained on Persona Chat, \textbf{KV-MemNN} -- Key Value Profile Memory Network, \textbf{Language Model} -- LSTM-based Language Model, \textbf{Seq2Seq}, and human annotated references -- \textbf{Human Generated Old}, and \textbf{Human Generated New}. For Topical-Chat (USR-TC and UniEval-TC), the systems are: Human annotations -- \textbf{Human Generated Old}, \textbf{Human Generated New}, and four systems that origin from Transformers with different decoding systems, such as \textbf{Nucleus Decoding $p=0.3$}, \textbf{Nucleus Decoding $p=0.5$}, \textbf{Nucleus Decoding $p=0.7$}, \textbf{Argmax Decoding} -- greedy decoding.

\paragraph{3. Neural Text Summarization} (TextSumm) \cite{grusky-etal-2018-newsroom,fabbri-etal-2021-summeval} focuses on a compression type of NLG where the main objective is to generate a concise version of texts, yet maintaining the salient information expressed in the document sources. The quality of system outputs is mainly assessed based on human evaluation aspects that fit into the objective of the task, such as \emph{coherence, consistency, fluency,} and \emph{relevance}. 

\paragraph{End-to-End NLG Systems} In \textbf{Newsroom} dataset \cite{grusky-etal-2018-newsroom}, the systems are divided into \textbf{Extractive} approach: 

\begin{itemize}
    \item \textbf{TextRank} \cite{mihalcea-tarau-2004-textrank} -- unsupervisedly rank sentences in document to form a summary with an approach similar to Google PageRank \cite{};
    \item \textbf{Extractive Oracle (Fragments)} -- Fragments $\mathcal{F}(A,S)$ are sets of shared sequences of tokens in $ A = \langle a_1, a_2, \ldots, a_n \rangle $ and $S = \langle s_1, s_2, \ldots, s_m \rangle $
\end{itemize}

\noindent \textbf{Abstractive} approach:

\begin{itemize}
    \item \textbf{Sequence-to-Sequence (Seq2Seq) / Attention}, Tensorflow implementation of \cite{rush-etal-2015-neural}\footnote{\url{https://modelzoo.co/model/textsum}}
\end{itemize}

 \noindent and \textbf{Mixed} approach:

 \begin{itemize}
     \item \textbf{Pointer Generator} \cite{see-etal-2017-get} with copying \cite{NIPS2015_29921001,gulcehre-etal-2016-pointing} and coverage \cite{tu-etal-2016-modeling} mechanism;
     \item \textbf{Lower Bound (Lede-3)} -- baseline approach, by copying the first sentence, first paragraph, or first $k$ words as the summary
 \end{itemize}

\noindent In \textbf{summEval} dataset, systems are divided into \textbf{Extractive}: 

\begin{itemize}
    \item \textbf{M1, NEUSUM} \cite{zhou-etal-2018-neural-document} -- scoring and selecting sentences based on hierarchical representation of a document;
    \item \textbf{M2, BanditSum} \cite{dong-etal-2018-banditsum} -- contextual bandit approach of summarization where the document is seen as context and the sequence of sentences to be included in the summary as action;
    \item \textbf{M3, LATENT} \cite{zhang-etal-2018-neural} -- views sentences in document as relevance binary labels of latent variables;
    \item \textbf{M4, REFRESH} \cite{narayan-etal-2018-ranking} -- a reinforcement approach by focusing on combining individually high-scoring sentences;
    \item \textbf{M5, RNES} \cite{10.5555/3504035.3504722} -- improving REINFORCE network by combining coherence model and ROUGE scores as a reward; 
    \item \textbf{M6, JECS} \cite{xu-durrett-2019-neural} -- scoring possible constituency-based compressed units;
    \item \textbf{M7, STRASS} \cite{bouscarrat-etal-2019-strass} -- selecting sentences based on the closest embeddings to the document embedding;
\end{itemize}

\noindent and \textbf{Abstractive}: 

\begin{itemize}
    \item \textbf{M8, Pointer Generator} \cite{see-etal-2017-get} -- encoder decoder model where the decoder can generate samples based on the log-likelihood of words in vocabulary or copy words from the sentence source;
    \item \textbf{M9, Fast-abs-rl} \cite{chen-bansal-2018-fast} -- improves Pointer Networks with ROUGE-L reward of REINFORCE;
    \item \textbf{M10, Bottom-up} \cite{gehrmann-etal-2018-bottom} -- decoding method with content selection model to restrict the copy attention distribution of pretrained Pointer Generation Network during inference;
    \item \textbf{M11, Improve-abs} \cite{kryscinski-etal-2018-improving} -- augments the decoder with external LSTM-based Language Model and RL-based objective;
    \item \textbf{M12, Unified-ext-abs} \cite{hsu-etal-2018-unified} -- aligns word-level attention scores of abstractive model with sentence level attention based on the probability outputs of extractive model;
    \item \textbf{M13, ROUGESal} \cite{pasunuru-bansal-2018-multi} -- improves reinforcement approach by using three types of rewards: keyphrase-based salience, entailment-based, and ROUGE-based reward;
    \item \textbf{M14, Multi-task (Ent+QG)} \cite{guo-etal-2018-soft} -- a multi-task learning approach with question and entailment generation as auxiliary tasks;
    \item \textbf{M15, Closed book decoder} \cite{jiang-bansal-2018-closed} -- introduces copy-less and attention-less decoder on Pointer Generator Network;
    \item \textbf{M16, SENECA} \cite{sharma-etal-2019-entity} -- combines entity-aware content selection module and abstractive generation module;
    \item \textbf{M17, T5} \cite{10.5555/3455716.3455856} -- improves Transformers-based architecture by exploring the limitation of various transfer learning approaches;
    \item \textbf{M18, NeuralTD} \cite{bohm-etal-2019-better} -- define RL-based reward function based on 2500 human evaluation outcomes ;
    \item \textbf{M19, BertSum-abs} \cite{liu-lapata-2019-text} -- extend BERT with document-level encoder;
    \item \textbf{M20, GPT-2} \cite{ziegler2019fine} -- finetune GPT-2 on human summaries with a reinforcement learning framework;
    \item \textbf{M21, UniLM} \cite{NEURIPS2019_c20bb2d9} -- use three language model tasks as pretrianing objective: unidirectional, bidirectional, and sequence-to-sequence prediction;
    \item \textbf{M22, BART} \cite{lewis-etal-2020-bart} -- use denoising autoencoder for pretraining sequence-to-sequence task;
    \item \textbf{M23, Pegasus} \cite{pmlr-v119-zhang20ae} -- model is trained on documents after removing important sentences.
\end{itemize}


\subsubsection{Types of automatic Metrics}
\label{sec:auto-metrics}

Figure~\ref{fig:auto_diag} shows the classification of metrics based on whether they are task-agnostic or human-aligned. We briefly discuss the categorization as follows:

\paragraph{Task-agnostic metrics}
Task-agnostic metric refers to a category of NLG evaluation metric that does not need task-specific design or contextual knowledge prior to its utilization in a new NLG task.

\begin{figure}[!t]
    \centering
    \includegraphics[width=\linewidth]{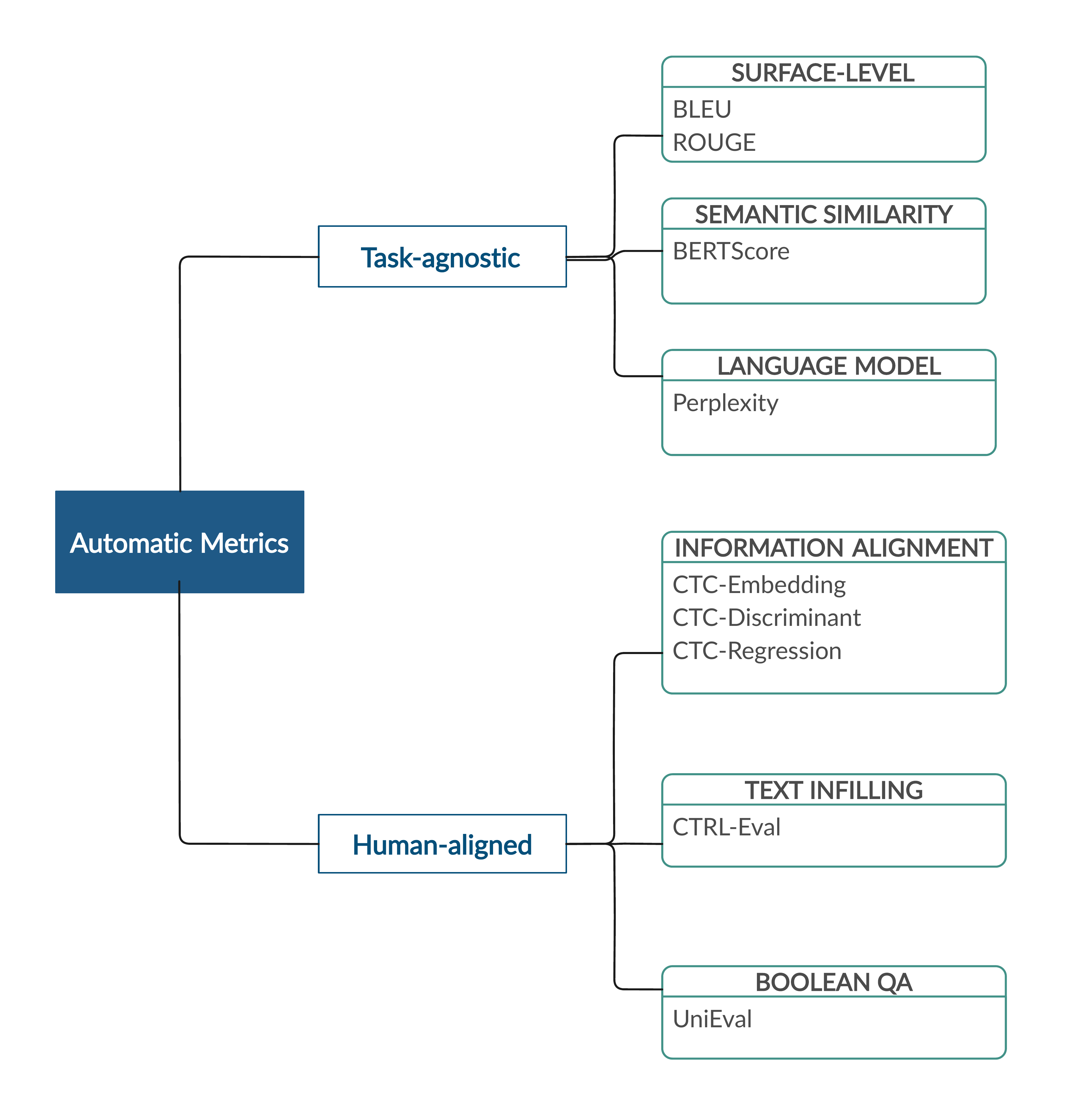}
    \caption{Automatic metrics in this study.}
    \label{fig:auto_diag}
\end{figure}

\begin{itemize}
    \item \textbf{Surface-level} refers to automatic metrics that mainly assess the quality of system outputs based on word-overlapping or string-based matching techniques between the generation outputs and human-generated references. Our study specifically focuses on two surface-level-based similarity metrics: Bilingual Evaluation Understudy (\textbf{BLEU}) \cite{papineni-etal-2002-bleu} -- computes $n$-gram precision of the generation outputs w.r.t. the corresponding ground truth references; Recall-Oriented Understudy for Gisting Evaluation (\textbf{ROUGE}) \cite{lin-2004-rouge} -- measures how good the system at recalling $n$-grams from human text references;
    \item \textbf{Semantic similarity} refers to metrics that measure the similarity between system outputs and text references based on the distance of textual features $\mathcal{X}$ in an embedding space $\mathcal{X} \in R$. In many cases, the mapping from texts to the corresponding vector representations $R$ requires a Deep Neural Network as an encoder, such as by utilizing pretrained Language Models (BERT) \cite{devlin-etal-2019-bert} or word embeddings \cite{10.5555/944919.944966,DBLP:journals/corr/abs-1301-3781,NIPS2013_9aa42b31}. In this study, we focus on investigating \textbf{BERTScore} \cite{Zhang*2020BERTScore:} to assess to what degree the generation outputs are similar to the given contexts (e.g. text sources, reference summaries, contextual knowledge, or control attributes);
    \item \textbf{Language Model-based metric} refers to evaluation metric that define the quality of generation outputs by linking the outputs to the surprisal score of an independent pre-trained Language Model -- where the surprisal of a word is mainly described as the negative logarithm of the word probability given preceding context words. \textbf{Perplexity} \cite{brown-etal-1992-estimate} is an example of automatic evalution metric that is defined based on the entropy of Language Model. Given machine-generated texts as the inputs of a pretrained LM (e.g. GPT-2), \textbf{Perplexity}  scores are the exponents of Negative Log-Likelihood (NLL) of the inputs;
\end{itemize}

\paragraph{Human-aligned metrics} refers to automatic metrics that translate multi-dimensional explainable human evaluation aspects (e.g. Coherence, Consistency) into measureable statistical features of an evaluation metric. We further classify human-aligned automatic metrics into two categories as follows:

\begin{itemize}
    \item \textbf{Single-aspect} views multi-dimensional human-like aspects or qualities as independent entities. 
    \begin{itemize}
        \item \textbf{CTC} \cite{deng-etal-2021-compression} -- is an automatic metric that the main objective is to \textbf{align information} between input, context, and output texts in \textbf{Compression}-based NLG (Summarization), \textbf{Transduction}-based NLG (Style Transfer), and \textbf{Creation}-based NLG (Dialogue Response Generation). The alignment function is estimated by \textbf{Embedding Matching ($E$), Discriminative Model ($D$), and Aggregated Regression ($R$)}. For example, in a compression task, \textbf{Consistency} aspect is described as the average of the alignment score ($f_E(.)$, $f_D(.)$, or $f_R(.)$) between the summarization outputs $y$ and the source $x$. Although \textbf{CTC} metric assesses the quality of system outputs based on multiple human evaluation aspects, the aspects are measured independently. Recent report \cite{} also discloses that \textbf{CTC} scores are bias to particular human-like aspect or quality. For example, \textbf{CTC-E Consistency} is highly correlated to consistency score based on human ratings, but it cannot explain the other human evaluation aspects. Therefore, our study classifies the metric as single-aspect human-aligned metric;
        \item \textbf{CtrlEval} \cite{ke-etal-2022-ctrleval} -- is unsupervised reference-less metric in Controlled Generation \cite{Dathathri2020Plug}. The metric translates three human evaluation aspects: Consistency, Coherence, Relevance into a \textbf{Text Infilling} objective. That is, given the input $I = (X, a, Y)$ consisting of prefix sentence $X$, control attribute $a$, and the generation output $Y$, the score is calculated by projecting pair of sequences from $I$ to $N$-number of pattern evaluators, where each pattern evaluator's score is estimated by the log probability outputs of pretrained model.;
    \end{itemize}
    \item \textbf{Multi-aspect} introduces a unifying perspective of multi-aspect human-like qualities via multi-task and continual learning objectives.
    \begin{itemize}
        \item \textbf{UniEval} \cite{zhong-etal-2022-towards} -- re-frames evaluation aspect as a Boolean Question Answering (QA) objective. For example, for a \textbf{Coherence} aspect, given a summarization output and the corresponding document source, the metric calculates the performance score by modeling a binary classification task (Yes/No) for a question ``\emph{Is this a coherent summary of the document?}''. Given $n$-multi dimensional aspects $d=(d_1, \ldots, d_n)$, the generation outputs $x$, reference texts $y$ (if applicable), and context $c$, the quality of the system outputs is measured based on the probability of the system generating words that can be either classified as positive and negative samples for addressing question $q_i$:
        
    \end{itemize}
\end{itemize}

\begin{equation}
        \begin{aligned}
        s_i = \frac{P(\textnormal{``Yes''}|x,y,c,q_i)}{P(\textnormal{``Yes''}|x,y,c,q_i) + P(\textnormal{``No''}|x,y,c,q_i)}
        \end{aligned}
    \vspace{-.5em}
    \label{eq:unieval}
    \end{equation}

\subsection{Assessment setups}

\paragraph{Data Preprocessing}

\begin{itemize}
    \item \textbf{summEval, Newsroom, UniEval-summ} (\textbf{TextSumm}) -- We use standard data preprocessing: we remove punctuation and non-textual (i.e. numeric and abbreviation) features; we also substitute latin abbreviation, such as \emph{i.e.} to \emph{id est} and \emph{e.g.} to \emph{exempli gratia}; prior to using the data to calculate the scores based on \textbf{Perplexity, CTC, CtrlEval,} and \textbf{UniEval} metrics. Specific to \textbf{CtrlEval}, we mainly utilize tf-idf weights in \cite{ke-etal-2022-ctrleval} \footnote{\url{https://github.com/thu-coai/CTRLEval}}, but we additionally generate relevant prompt and verbal dictionary for the summarization task. as shown in Table~\ref{tab:CtrlEval-prompts}.
    \item \textbf{USR-PC, USR-TC, UniEval-TC} (\textbf{DiagGen}) -- Specific to \textbf{CTC}-based evaluator, the format of references (list of personas) as relevance-based attribute is adjusted accordingly to follow the input-output structure of the pretrained evaluator. That is by transforming line-separable personas into a single line of text input separated by a character ``||''.
    \item \textbf{UBER-Topic, CTRL-Topic, CtrlEval-Topic} (\textbf{CtrlGen}) -- Data preprocessing follows the procedur in Text Summarization task. Since the nature of benchmark datasets in Controlled Generation is reference-less and open-endedness - no human-generated texts as ground truth references, we use the concatenation between control attribute (topic category, such as ``Science'') and its corresponding list of relevant keywords as a means of reference.
\end{itemize}

\begin{table*}[!ht]
\resizebox{.995\textwidth}{!}{
    \centering
    \begin{tabular}{ l l p{7.5cm} p{4.5cm}}
     \hline
     \thickhline
    \bf NLG Task & \bf Benchmark dataset & \bf Prompts & \bf Verbal Dict. \\
     \hline
     \thickhline
      TextSumm   & summEval, Newsroom & $\langle$ \emph{gen\_result} $\rangle$ Article: $\langle$ \emph{mask\_token} $\rangle$  & N/A \\
      & & $\langle$ \emph{gen\_result} $\rangle$ Summary:  $\langle$ \emph{mask\_token} $\rangle$  & N/A \\
      & & $\langle$ \emph{gen\_result} $\rangle$ It was about  $\langle$ \emph{mask\_token} $\rangle$  & N/A \\
      \hline
      DiagGen   & USR-PC & $\langle$ \emph{gen\_result} $\rangle$ Persona: $\langle$ \emph{mask\_token} $\rangle$ & list of system's and human evaluator's personas\\
       &  & The persona of $\langle$ \emph{gen\_result} $\rangle$ is $\langle$ \emph{mask\_token} $\rangle$  &  \\
       &  & $\langle$ \emph{gen\_result} $\rangle$ contains  $\langle$ \emph{mask\_token} $\rangle$ persona &  \\
         & USR-TC, UniEval-TC & $\langle$ \emph{gen\_result} $\rangle$ It was about  $\langle$ \emph{mask\_token} $\rangle$ & context \\
         & & $\langle$ \emph{gen\_result} $\rangle$ It was related to  $\langle$ \emph{mask\_token} $\rangle$ &  \\
        \hline
     CtrlGen & UBER-Topic, CTRL-Topic & $\langle$ \emph{gen\_result} $\rangle$ News:  $\langle$ \emph{mask\_token} $\rangle$  & computers, politics, religion, \\
     & & $\langle$ \emph{gen\_result} $\rangle$ It was about  $\langle$ \emph{mask\_token} $\rangle$  &   science, legal, clickbait, space, military\\
     \hline
     \thickhline
    \end{tabular}
    }
    \caption{Examples of prompts and verbal dictionary as auxiliary inputs for CtrlEval metric.}
    \label{tab:CtrlEval-prompts}
\end{table*}

\paragraph{References and Human-like Aspects} Our study uses the following frame of references, which are dependent to the target NLG evaluation task or benchmark dataset and the characteristic of automatic metrics:

\begin{itemize}
    \item \textbf{summEval} (\textbf{TextSumm}) -- The dataset uses $n$-references ($n=11$) as ground truth human-generated summaries. For each system output and the corresponding references, the score based on \textbf{BLEU},  \textbf{ROUGE}, \textbf{BERTScore}, and human ratings (\textbf{Coherence, Consistency, Fluency, Relevance}) are already included in dataset. For \textbf{BLEU},  \textbf{ROUGE}, and \textbf{BERTScore}, we average the metric scores based on 1-reference and 11-references. 
    
    Our work additionally compute the scores based on \textbf{Perplexity, CTC, CtrlEval,} and \textbf{UniEval} metrics. \textbf{Perplexity} mainly uses the system's outputs as the input $x$ of the metric. For \textbf{CTC}, we use 1-reference only as the ground truth target and average the scores based on embedding-based CTC (CTC-E), discriminator-based CTC (CTC-D), and regressor-based CTC (CTC-R) w.r.t. the two aspects of evaluation: \textbf{``Consistency''} and \textbf{``Relevance''}. The inputs for CTC metric are $x=\{\textnormal{\emph{docs, hypos, refs}}\}$ -- where $\textnormal{\emph{docs}}$ denotes document source to be summarized, $\textnormal{\emph{hypos}}$ denotes the system's generation outputs, and $\textnormal{\emph{refs}}$ is ground truth human-generated summaries. 
    
    For \textbf{CtrlEval} and \textbf{UniEval}, we use $11$-references as evaluation target for the metrics. For \textbf{CtrlEval}, the performance score is computed based on \textbf{``Coherence''} aspect by solely utilizing the system outputs as the input sources for pretrained GPT-2. 
    
    For \textbf{UniEval}, the evaluator is pretrained on summarization task for assessing four aspects: \textbf{``Coherence''}, \textbf{``Consistency''}, \textbf{``Fluency''}, and \textbf{``Relevance''}. For assessing ``Coherence'' and ``Consistency'' aspects, UniEval uses document source and the system outputs as the inputs for pretrained evaluator. The system outputs is used solely as inputs for measuring ``Fluency'', while the generation outputs and ground truth references are compared for measuring ``Relevance'' aspect. 
    
    \item \textbf{Newsroom} (\textbf{TextSumm}) -- The evaluation setup for Newsroom dataset is similar to summEval, except that Newsroom does not include ground truth human references. Instead, the title of articles is used as a means of reference for assessing the quality of system outputs.
    
    \item \textbf{UniEval-summ} (\textbf{TextSumm}) -- is a subset of summEval. Therefore, the evaluation setup follows the configuration in summEval data.
    
    \item \textbf{USR-PC} (\textbf{DiagGen}) -- is composed of three source of textual inputs for the evaluation metrics: persona of the model (NLG system) and human evaluators as a background knowledge (fact), the preceding dialogue as a context, and the system responses (generation outputs). 

    \textbf{BLEU, ROUGE} are computed by comparing between the system responses and the concatenation of document source and factual or contextual knowledge (i.e. list of personas in USR-PC and document title in USR-TC). While, \textbf{BERTScore}is computed by comparing between system's responses and document sources.

   \textbf{CTC} scores are measured based on ``Engagingness'' and ``Groundedness'' (Use Knowledge) aspects, two aspects out of total five aspects based on human ratings (Understandable, Natural, Maintains Context, Engaging, Use Knowledge). CTC-based engagingness is measured by utilizing (i) the concatenated version of factual knowledge (personas) and dialogue history, and (ii) system responses as inputs to be compared. While, CTC-based groundedness measures the relevance of information by inspecting how the system responses comply with the predefined factual knowledge.

    \textbf{CtrlEval} scores are measured based on ``Coherence'', ``Consistency'', and ``Relevance'' aspects. CtrlEval-Coherence uses the concatenation of dialogue history and system response as input. CtrlEval-Consistency measures how consistent the system response w.r.t. the prefix or dialogue history. While, CtrlEval-Relevance compares the degree of relevance between the generated responses and the predefined personas.

   \textbf{UniEval} scores are computed based on human evaluation aspects included in \textbf{USR-PC} data: UnieEval-Understandability, UniEval-Naturalness, UniEval-Coherence, UniEval-Engagingness, UniEval-Groundedness, and UniEVal-Overall; given dialogue histories as source, list of personas as contextual knowledge, and the system responses as output to be evaluated. 
    
    \item \textbf{USR-TC, UniEval-TC} (\textbf{DiagGen}) -- The main difference between USR-TC and USR-PC is that the two benchmarks use different factual knowledge as a means of reference for model or metric. In USR-PC, the reference is the predefined list of model and human personas as multi-turn agents in a dialogue system. While, in USR-TC, the predefined knowledge-grounded conversation is used as a means of reference for evaluating systems and metrics in this study.
    
    \item \textbf{UBER-Topic, CTRL-Topic, CtrlEval-Topic} (\textbf{CtrlGen}) -- are mainly composed of prefixes, the perturbed version of generation outputs, and control attributes (i.e. topic categories) as textual inputs for the evaluation metrics. The contextual knowledge is constructed by concatenating topic category as control attribute for each prefix sample and the corresponding list of keywords as a pointer to particular topic or domain.

    \textbf{BERTScore} is defined based on the comparison between the system's generated outputs and the control attributes as contextual knowledge. For each system output, we construct the context by concatenating topic category (e.g. ``Science'') and its corresponding list of relevant keywords. While, \textbf{Perplexity} is measured by projecting the system outputs as inputs for pretrained GPT-2.

    \textbf{CTC} measures two aspects: Consistency and Relevance. We specifically use ``SummarizationScorer'' of CTC for assessing the quality of system outputs in Controlled Generation task because the task share more similar characteristic to Text Summarization than task in Dialogue Response Generation. The setup follows the configuration of Summarization-based CTC evaluation.

    \textbf{CtrlEval} measures three evaluation aspects: Coherence, Consistency, and Relevance. CtrlEval-Coherence outputs the pattern evaluator score by pairing sentences in the generation outputs as a text infilling task. CtrlEval-Consistency uses prefixes and system outputs as the inputs of the metric. While, CtrlEval-Relevance measures whether the generation outputs are relevant to the given control attributes (topic categories).  

    \textbf{UniEval} measures four aspects: Coherence, Consistency, Fluency, and Relevance. The setup follows the configuration of summarization-based UniEval evaluation, but the reference list is defined based on the concatenation between control attribute (topic category) and its corresponding pointer words (keywords). 
    
\end{itemize}

\subsection{Experiment Results}

\subsubsection{Transfer Experiment}

Table~\ref{tab:corr-score-id-ood}-~\ref{tab:corr-score-nlg} shows the correlation score between automatic metrics and human ratings across NLG tasks (ID and OOD).

\begin{table}[!ht]
    \centering
    \resizebox{.45\textwidth}{!}{
    \begin{tabular}{l c c c}
     \hline
     \thickhline
      \bf Automatic metrics   & \bf  ID & \bf  Semantic-Shift & \bf  Domain-Shift \\
       \hline
     \thickhline
       LM-Perplexity  & 0.170 & 0.022 & -0.116\\
       Surface-level (BLEU \& ROUGE)  & 0.215 & 0.193 & 0.000 \\
       Semantic (BERTScore)  & 0.213 & 0.075 & 0.054 \\
       Single-CTC  & 0.259 & 0.091 & 0.024 \\
       Single-CTRLEval  & 0.145 & 0.156 & 0.058 \\
       Multi-UniEval  & 0.445 & 0.257 & 0.006 \\
        \hline
     \thickhline
    \end{tabular}
    }
    \caption{Correlation level to human scores across ID and OOD samples}
    \label{tab:corr-score-id-ood}
\end{table}

\begin{table}[!ht]
    \centering
    \resizebox{.45\textwidth}{!}{
    \begin{tabular}{l c c c}
     \hline
     \thickhline
     \bf  Automatic metrics   & \bf  TextSumm & \bf  DiagGen & \bf  CtrlGen \\
       \hline
     \thickhline
       LM-Perplexity  & -0.116 & 0.170 &  0.022 \\
       Surface-level (BLEU \& ROUGE)  & 0.215 & 0.193 & 0.000 \\
       Semantic (BERTScore)  & 0.213 & 0.074 & 0.054 \\
       Single-CTC  & 0.026 & 0.147 & 0.024 \\
       Single-CTRLEval  & 0.156 & 0.074 & 0.086  \\
       Multi-UniEval  & 0.341 & 0.298 & 0.006 \\
        \hline
     \thickhline
    \end{tabular}
    }
    \caption{Correlation level to human scores across NLG tasks}
    \label{tab:corr-score-nlg}
\end{table}

\paragraph{Sample Analysis} In this section, we sample data in In-Domain (ID) and Out-of-Domain subsets to further analyze the contexts in which automatic metrics are not in alignment with human judgments. The samples are mainly grouped based on the agreement-level of multi-aspect human ratings (low vs. high) across ID and OOD subsets (Figure~\ref{fig:transfer-ood}) and NLG use cases (Figure~\ref{fig:transfer-task}).

\begin{table*}[!ht]
    \centering
    \resizebox{.99\textwidth}{!}{
    \begin{tabular}{ l p{5.0cm} p{2.5cm} c c c c p{2.75cm} p{2.3cm} p{2.75cm}}
     \hline
     \thickhline
      \bf System & \bf System Outputs &  \bf Human Rating & \multicolumn{7}{c}{\textbf{Metric Score}} \\
      & &  &  \bf Perplexity $\downarrow$ & \bf BLEU (\%) $\uparrow$ & \bf ROUGE $\uparrow$  & \bf BERTScore $\uparrow$  & \bf CTC $\uparrow$ & \bf CtrlEval $\uparrow$ & \bf UniEval $\uparrow$ \\
    \hline
     M12 & {\small{paul merson has restarted his row with andros townsend after the tottenham midfielder was brought on with only seven minutes remaining in his team 's 0-0 draw with burnley . merson initially angered townsend for writing in his sky sports column that ` if andros townsend can get in -lrb- the england team -rrb- then it opens it up to anybody . ' paul merson had another dig at andros townsend after his appearance for tottenham against burnley .}} & {\small{Coherence: 5, Consistency: 5, Fluency: 5, Relevance: 4, Average: 4.75}} & 78.49 & 10.131 & 0.283 & 0.422 & {\small{E-Consistency: 0.882, E-Relevance: 0.548, D-Consistency: 0.950, D-Relevance: 0.579, R-Consistency: 0.939, R-Relevance: 0.557}} & {\small{Coherence: (-)3.594}} & {\small{Coherence: 0.860, Consistency: 0.784, Fluency: 0.648, Relevance: 0.207 }}\\
     &      &  & &&& &&&\\
     M23 & {\small{paul merson had a dig at andros townsend after his appearance for tottenham . townsend was brought on in the 83rd minute for tottenham against burnley . \'just been watching the game, did you miss the coach? \#rubberdub \#7minutes,\' merson wrote on twitter .}} & {\small{Coherence: 5, Consistency: 5, Fluency: 5, Relevance: 5, Average: 5}} & 131.58 & 6.028 & 0.308 & 0.324 & {\small{E-Consistency: 0.896 , E-Relevance: 0.566, D-Consistency: 0.959, D-Relevance: 0.559, R-Consistency: 1.053, R-Relevance: 0.624}} & {\small{Coherence: (-)3.200}} & {\small{Coherence: 0.929, Consistency: 0.933, Fluency: 0.881, Relevance: 0.878 }}\\
     &      &  & &&& &&&\\
     M11 & {\small{paul merson was brought on with only seven minutes remaining in his team 's 0-0 draw with burnley . andros townsend scored the tottenham midfielder in the 89th minute . paul merson had another dig at andros townsend after his appearance . the midfielder had been brought on to the england squad last week . click here for all the latest arsenal news news.}}     & {\small{Coherence: 1, Consistency: 1, Fluency: 2, Relevance: 1, Average: 1.25}} & 70.055 & 11.912 & 0.310 & 0.399 & {\small{E-Consistency: 0.859, E-Relevance: 0.535, D-Consistency: 0.773, D-Relevance: 0.481, R-Consistency: 0.793, R-Relevance: 0.491}} & {\small{Coherence: (-)2.869}} & {\small{Coherence: 0.103, Consistency: 0.542, Fluency: 0.589, Relevance: 0.122 }} \\
      &     &  & &&& &&&\\
    \hline 
    \multicolumn{10}{p{28.cm}}{\textbf{Source: } {\small{Paul Merson has restarted his row with Andros Townsend after the Tottenham midfielder was brought on with only seven minutes remaining in his team's 0-0 draw with Burnley on Sunday. 'Just been watching the game, did you miss the coach? \#RubberDub \#7minutes,' Merson put on Twitter. Merson initially angered Townsend for writing in his Sky Sports column that 'if Andros Townsend can get in (the England team) then it opens it up to anybody.' Paul Merson had another dig at Andros Townsend after his appearance for Tottenham against Burnley Townsend was brought on in the 83rd minute for Tottenham as they drew 0-0 against Burnley Andros Townsend scores England's equaliser in their 1-1 friendly draw with Italy in Turin on Tuesday night The former Arsenal man was proven wrong when Townsend hit a stunning equaliser for England against Italy and he duly admitted his mistake. 'It's not as though I was watching hoping he wouldn't score for England, I'm genuinely pleased for him and fair play to him â€“ it was a great goal,' Merson said. 'It's just a matter of opinion, and my opinion was that he got pulled off after half an hour at Manchester United in front of Roy Hodgson, so he shouldn't have been in the squad. 'When I'm wrong, I hold my hands up. I don't have a problem with doing that - I'll always be the first to admit when I'm wrong.' Townsend hit back at Merson on Twitter after scoring for England against Italy Sky Sports pundit  Merson (centre) criticised Townsend's call-up to the England squad last week Townsend hit back at Merson after netting for England in Turin on Wednesday, saying 'Not bad for a player that should be 'nowhere near the squad' ay \@PaulMerse?' Any bad feeling between the pair seemed to have passed but Merson was unable to resist having another dig at Townsend after Tottenham drew at Turf Moor.}}} \\
   & &  & &&& &&&\\
    \hline 
    \multicolumn{10}{p{28.cm}}{\textbf{$1^{st}$ Reference: } {\small{Andros Townsend an 83rd minute sub in Tottenham's draw with Burnley. He was unable to find a winner as the game ended without a goal. Townsend had clashed with Paul Merson last week over England call-up.}}} \\
    &&  & &&& &&&\\
    \hline
     \multicolumn{10}{p{28.cm}}{\textbf{$2^{nd}$ Reference: } {\small{Sports columnist Paul Merson and Andros Townsend are in the midst of a twitter feud. Merson started it when Townsend was called up and wrote something disparaging about him in his column. Since then things have gone back and forth between the two.}}} \\
    &  & &&& &&&\\
    \hline
     \multicolumn{10}{p{28.cm}}{\textbf{$3^{rd}$ Reference: } {\small{Merson is angered by the statement made by Townsend in his Sky Sports column. Merson threw a dig at Townsend after scoring his last game.}}} \\
    &&  & &&& &&&\\
     \hline
     \thickhline
    \end{tabular}
    }
    \caption{The system outputs in \textbf{summEval} with high agreement level between multiple human-like aspects for high human ratings ($N$-sample $=1987 (39\%)$) and low human ratings ($N$-sample $=43 (0.8\%)$). BLEU score is by default represented as percentage rather than decimal in benchmark dataset. Both BLEU and ROUGE scores are based on an averaged between 1-reference score and 11-references score.}
    \label{tab:sample1}
\end{table*}

\begin{table*}[!ht]
    \centering
    \resizebox{.99\textwidth}{!}{
    \begin{tabular}{ l p{5.0cm} p{2.5cm} c c c c p{2.75cm} p{2.3cm} p{2.75cm}}
     \hline
     \thickhline
      \bf System & \bf System Outputs &  \bf Human Rating & \multicolumn{7}{c}{\textbf{Metric Score}} \\
      & &  &  \bf Perplexity $\downarrow$ & \bf BLEU (\%) $\uparrow$ & \bf ROUGE $\uparrow$  & \bf BERTScore $\uparrow$  & \bf CTC $\uparrow$ & \bf CtrlEval $\uparrow$ & \bf UniEval $\uparrow$ \\
     \hline
     \thickhline
     M20 & {\small{Varvara traveled 14,000 miles across the Pacific Ocean. (Hat tip: The Daily Beast)}} & {\small{Coherence: 4, Consistency: 2, Fluency: 5, Relevance: 1, Average: 3}} & 35.68 & 4.17 & 0.204 & 0.285 & {\small{E-Consistency: 0.848, E-Relevance: 0.518, D-Consistency: 0.766, D-Relevance: 0.348, R-Consistency: 0.645, R-Relevance: 0.322}} & {\small{Coherence: (-)4.464}} & {\small{Coherence: 0.113, Consistency: 0.721, Fluency: 0.945, Relevance: 0.789 }}\\
      &     &  & &&& &&&\\
     M8 & {\small{the whale , named varvara , swam nearly 14,000 miles ( 22,500 kilometers ) . it said the previous record was set by a humpback whale that swam a mere 10,190-mile round trip between the `` warm breeding waters of the arctic and antarctic regions '' .}}     & {\small{Coherence: 2, Consistency: 4, Fluency: 5, Relevance: 2, Average: 3.25}} & 50.71 & 28.74 & 0.443 & 0.613 & {\small{E-Consistency: 0.908, E-Relevance: 0.571, D-Consistency: 0.951, D-Relevance: 0.627, R-Consistency: 0.970, R-Relevance: 0.653}} & {\small{Coherence: (-)3.228}} & {\small{Coherence: 0.682, Consistency: 0.957, Fluency: 0.690, Relevance: 0.112}} \\
     &      &  & &&& &&&\\
    \hline 
    \multicolumn{10}{p{28.cm}}{\textbf{Source: } {\small{(CNN)A North Pacific gray whale has earned a spot in the record books after completing the longest migration of a mammal ever recorded. The whale, named Varvara, swam nearly 14,000 miles (22,500 kilometers), according to a release from Oregon State University, whose scientists helped conduct the whale-tracking study. Varvara, which is Russian for "Barbara," left her primary feeding ground off Russia's Sakhalin Island to cross the  Pacific Ocean and down the West Coast of the United States to Baja, Mexico. Varvara's journey surpassed a record listed on the Guinness Worlds Records website. It said the previous record was set by a humpback whale that swam a mere 10,190-mile round trip between the "warm breeding waters near the equator and the colder food-rich waters of the Arctic and Antarctic regions." Records are nice, but Bruce Mate, the lead author of the study, thinks the long trip might say more about the whale than just its ability to swim. During her 14,000-mile journey, Varvara visited "three major breeding areas for eastern gray whales," which was a surprise to Mate, who is also the director of the Marine Mammal Institute at Oregon State University. "For her to go to Mexico," Mate said, "It's pretty strong evidence that it's where she's from." Varvara was thought to be an endangered western whale, but her ability to "navigate across open water over tremendously long distances is impressive," he said in the release, which could mean that some western gray whales are actually eastern grays. With only 150 western gray whales believed to be in existence, that number might be even lower. "Past studies have indicated genetic differentiation between the species, but this suggests we may need to take a closer look," Mate said. Fourth baby orca born this season}}} \\
    &&  & &&& &&&\\
    \hline 
    \multicolumn{10}{p{28.cm}}{\textbf{$1^{st}$ Reference: } {\small{The whale, Varvara, swam a round trip from Russia to Mexico, nearly 14,000 miles. The previous record was set by a humpback whale that migrated more than 10,000 miles.}}} \\
   & &  & &&& &&&\\
    \hline
     \multicolumn{10}{p{28.cm}}{\textbf{$2^{nd}$ Reference: } {\small{A record for the longest distance migration of a mammal was shattered recently by a north pacific gray whale. The whale made a trip of 14,000 miles.}}} \\
   & &  & &&& &&&\\
    \hline
     \multicolumn{10}{p{28.cm}}{\textbf{$3^{rd}$ Reference: } {\small{The longest mammalian migration was just recorded by a pacific gray whale. It swam over 14,000 miles in the process. There are only about 150 gray whales known.}}} \\
    & &  & &&& &&&\\
    \hline
     M11 & {\small{jordan henderson is set to sign a new long-term contract at anfield . the club \'s vice-captain had 14 months remaining on his current contract . henderson is the third major player in liverpool \'s fa cup . the fa cup fourth round . raheem sterling is expected to return to liverpool in the summer .}} & {\small{Coherence: 1, Consistency: 4, Fluency: 1, Relevance: 4, Average: 2.5}} & 45.03 & 28.72 & 0.410 & 0.589 & {\small{E-Consistency: 0.868, E-Relevance: 0.546, D-Consistency: 0.803, D-Relevance: 0.538, R-Consistency: 0.834, R-Relevance: 0.517}} & {\small{Coherence: (-)2.635}} & {\small{Coherence: 0.018, Consistency: 0.637, Fluency: 0.675, Relevance: 0.011 }}\\
      &     &  & &&& &&&\\
    M8 & {\small{jordan henderson has provided liverpool with a lift after their fa cup heartache . the club \'s vice-captain had 14 months remaining on his current contract . his advisors had been in talks with liverpool since the beginning of this season .}}     & {\small{Coherence: 1, Consistency: 5, Fluency: 5, Relevance: 2, Average: 3.25}} & 68.84  & 21.68 & 0.403 & 0.498 & {\small{E-Consistency: 0.922, E-Relevance: 0.581, D-Consistency: 0.983, D-Relevance: 0.642, R-Consistency: 1.066, R-Relevance: 0.622}} & {\small{Coherence: (-)4.360}} & {\small{Coherence: 0.973, Consistency: 0.939, Fluency: 0.639, Relevance: 0.711}} \\
     &      &  & &&& &&&\\
    \hline 
    \multicolumn{10}{p{28.cm}}{\textbf{Source: } {\small{Jordan Henderson has provided Liverpool with a lift after their FA Cup heartache by agreeing a new long-term contract. The club's vice-captain had 14 months remaining on his current contract and his advisors had been in talks with Liverpool since the beginning of this season. They have now reached a resolution and Henderson is expected to put pen-to-paper on improved terms that are likely be worth in the region of Â£100,000. His new deal will run to 2020. Liverpool midfielder Jordan Henderson is set to sign a new long-term contract at Anfield Henderson chases down Aston Villa's Jack Grealish during Liverpool's FA Cup semi-final defeat at Wembley Henderson's new deal is worth around Â£100,000-a-week and will run until the summer of 2020 Henderson, 24, is the third big player in Brendan Rodgers' squad to agree a contract extension, following on from Daniel Sturridge and Philippe Coutinho. The England international, who was signed by Kenny Dalglish in June 2011 for Â£16million from Sunderland, has been one of the most improved players under Rodgers' watch. His form this season has been excellent and he has contributed 13 assists as well as seven goals from midfield; he will be considered for the role of club captain when Steven Gerrard moves to LA Galaxy. Talks with Raheem Sterling are not expected to resume until the end of the season but Ian Ayre, Liverpool's Chief Executive, last week said he expected the England forward to be at Anfield for 'a long time'. Henderson could replace Steven Gerrard as Liverpool captain when the 34-year-old departs this summer Liverpool boss Brendan Rodgers (right) is keen to tie-down Henderson with up to 10 players set to leave Raheem Sterling has rejected a new deal at Liverpool but talks are expected to resume in the summer}}} \\
    &&  & &&& &&&\\
    \hline 
    \multicolumn{10}{p{28.cm}}{\textbf{$1^{st}$ Reference: } {\small{Jordan Henderson is set to sign an improved deal with Liverpool. The 24-year-old midfielder has 14 months left on his current contract. Henderson could replace Steven Gerrard as club captain this summer. Liverpool will resume talks with Raheem Sterling at the end of the season.}}} \\
   & &  & &&& &&&\\
    \hline
     \multicolumn{10}{p{28.cm}}{\textbf{$2^{nd}$ Reference: } {\small{A player has signed onto a new contract with another team which is set to start in 2020. The player has shown to be quite impressive over the years and replaced a veteran last year.}}} \\
   & &  & &&& &&&\\
    \hline
     \multicolumn{10}{p{28.cm}}{\textbf{$3^{rd}$ Reference: } {\small{Jordan Henderson was heroic for Liverpool with a newly-signed contract. He has improved immensely over the years. He could very well replace Gerrard as team captain soon.}}} \\
    & &  & &&& &&&\\
     \hline
     \thickhline
    \end{tabular}
    }
    \caption{The system outputs in \textbf{summEval} with low agreement level between multiple human-like aspects. }
    \label{tab:sample1}
\end{table*}

\subsubsection{Aspect-level Evaluation}

Figure~\ref{fig:ks-diag-aspect} shows Kolmogorov-Smirnov (KS) scores for aspect-level evaluation in Dialogue Response Generation (DiagGen) and the corresponding similarity score to human preference.

\subsubsection{System-level Evaluation}

Table~\ref{tab:ks-textsumm}-\ref{tab:ks-ctrlgen} show Kolmogorov-Smirnov (KS) scores of both human and automatic metrics as a measure of metric's capability at distinguishing performance differences between independent NLG systems. Table~\ref{tab:pref-sim-textsumm}-\ref{tab:pref-sim-ctrlgen} show the preference similarity between human and automatic metrics at deciding the performance rank of the systems.

\subsection{Packages}
We use publicly available Python Packages for running the experiments, as listed in Table~\ref{tab:packages}. The prerequisite installation is provided in the shared implementation code.

\subsection{Hyperparameters}

\paragraph{BLEU} \textbf{Package}: evaluate, \url{https://huggingface.co/spaces/evaluate-metric/bleu}. \textbf{Parameters}: 'brevity\_penalty': 1.0 (default).

\paragraph{ROUGE} \textbf{Package}: evaluate, \url{https://huggingface.co/spaces/evaluate-metric/rouge}. 

\paragraph{BERTScore} \textbf{Package}: evaluate, \url{https://huggingface.co/spaces/evaluate-metric/bertscore}. \textbf{Model}: ``roberta-large\_L17\_no-idf\_version=0.3.12(hug\_trans=4.25.1)''.

\paragraph{Perplexity} \textbf{Package}: evaluate, \url{https://huggingface.co/spaces/evaluate-metric/perplexity}. \textbf{Model}: ``gpt2''.

\paragraph{CTC} \textbf{Package}: CTC. For Embedding-based alignment (CTC-E), we use BERTAligner/BERT embedding (default). For discriminative alignment (CTC-D), we use ``roberta-large''. For regressive alignment (CTC-R), we use BLEURTAligner.

\paragraph{CtrlEval} \textbf{Package}: CtrlEval. \textbf{Model}: ``google/pegasus-large''. We use default configuration in \url{https://github.com/thu-coai/CTRLEval}. We reuse the TfIdf features of the original work. For the other required external knowledge (prompt and verbal list), we adjust accordingly to the objective of target NLG task. The prompt and verbal files are provided in the shared data and code implementation.

\paragraph{UniEval} \textbf{Package}: UniEval. We use two types of pretrained evaluators in \url{https://github.com/maszhongming/UniEval}: UniEval-sum and UniEval-dialog. We re-use the multi-dimensional human evaluation aspects of the corresponding pretrained evaluators. We adjust the configuration of inputs-outputs of the evaluators based on the target NLG tasks.

\subsection{Computing Resources} 
Experiments were done in computing nodes of a HPC cluster with specifications of 4 GPUs Nvidia Tesla V100 (16GB RAM, 2560 tensor cores, 10480 CUDA cores, compute capability 7.0). 1 CPU Intel Xeon E5-2698v4 @ 2.2GHz (40 hyperthreads, RAM: 256GB). 

\newpage 

\begin{table}[!ht]
    \centering
    \resizebox{.45\textwidth}{!}{
    \begin{tabular}{l p{3.cm} p{5.5cm} }
     \hline
     \thickhline
      \bf Package name   &  \bf Version & \bf Link  \\
       \hline
     \thickhline
     Python  & 3.7.12 &  conda install \\
     Numpy  & 1.21.6 & pip install  \\
     Pandas  & 1.3.5 & pip install \\
      Matplotlib  & 3.5.2 & pip install \\
      NLTK  & 3.7  & pip install \\
      Pytorch  & 1.11.0+cu102 & conda install \\
      Transformers  & 4.25.1 & pip install \\
      Evaluate   & 0.2.2 & \url{https://github.com/huggingface/evaluate.git} \\
      CTC   & N/A & \url{https://github.com/tanyuqian/ctc-gen-eval.git}  \\
      CtrlEval   & N/A & \url{https://github.com/thu-coai/CTRLEval.git} \\
      UniEval  & N/A & \url{https://github.com/maszhongming/UniEval.git}  \\
         \hline
     \thickhline
    \end{tabular}
    }
    \caption{Python packages used in this study.}
    \label{tab:packages}
\end{table}

\begin{table}[!ht]
    \centering
    \resizebox{.45\textwidth}{!}{
    \begin{tabular}{l p{3.5cm} p{3.5cm}} 
    \hline
     \thickhline
      \bf Benchmark   &  \bf Easy pair & \bf Hard pair \\
       \hline
     \thickhline
     UBER-Topic & ('BR', 'BCR') & ('BC', 'BCR') \\
     & ('BC', 'BR') & ('B', 'BR') \\
     \hline
     CTRL-Topic & ('BCR', 'CTRL') & ('CTRL', 'WD')\\
     & ('BCR', 'WD') & \\
         \hline
     \thickhline
    \end{tabular}
    }
    \caption{System pairs in CtrlGen.}
    \label{tab:syspairs-ctrlgen}
\end{table}

\begin{table}[!ht]
    \centering

    \resizebox{.45\textwidth}{!}{
    \begin{tabular}{l p{3.75cm} p{3.75cm}}
     \hline
     \thickhline
      \bf Benchmark   &  \bf Easy pair & \bf Hard pair \\
       \hline
     \thickhline
     UniEval-summ & ('M11', 'M22') & ('M11', 'M9') \\
     & ('M11', 'M23') &  ('M13', 'M12') \\
     & ('M9', 'M22') & ('M23', 'M22') \\
     & ('M9', 'M23') & ('M11', 'M20') \\
     & ('M11', 'M2') & ('M17', 'M15') \\
     & ('M11', 'M0') & ('M0', 'M2') \\
     & ('M20', 'M2') & ('M2', 'M12') \\
     & ('M20', 'M0') & ('M17', 'M0')  \\
     & ('M11', 'M17')& ('M1', 'M13') \\
     & ('M20', 'M17') & ('M22', 'M23') \\
     & ('M20', 'M23') & ('M0', 'M22') \\
     & ('M20', 'M22') & \\
     \hline
     \thickhline
    \end{tabular}
    }
    \caption{System pairs in TextSumm (UniEval-Summ).}
    \label{tab:syspairs-summ1}
    
\end{table}

\begin{table}[!ht]
    \centering

    \resizebox{.45\textwidth}{!}{
    \begin{tabular}{l p{3.75cm} p{3.75cm}}
     \hline
     \thickhline
      \bf Benchmark   &  \bf Easy pair & \bf Hard pair \\
       \hline
     \thickhline
     summEval & ('M11', 'M22') & ('M11', 'M9') \\
     & ('M11', 'M23') & ('M13', 'M12') \\
     & ('M9', 'M22') & ('M23', 'M22') \\
     & ('M9', 'M23') & ('M11', 'M20') \\
     & ('M11', 'M2') & ('M23', 'M17') \\
     & ('M11', 'M0') & ('M0', 'M2') \\
     & ('M20', 'M2') & ('M5', 'M2') \\
     & ('M20', 'M0') & ('M17', 'M0') \\
     & ('M11', 'M17') &  ('M1', 'M13') \\
     & ('M20', 'M17') & ('M23', 'M23\_dynamicmix') \\
     & ('M11', 'M23\_dynamicmix') & \\
     & ('M20', 'M23\_dynamicmix') & \\
     & ('M20', 'M23') & \\
     & ('M20', 'M22') & \\
     \hline
     \thickhline
    \end{tabular}
    }
    \caption{System pairs in TextSumm (summEval).}
    \label{tab:syspairs-summ2}
    
\end{table}

\begin{table}[!ht]
    \centering

    \resizebox{.45\textwidth}{!}{
    \begin{tabular}{l p{3.75cm} p{3.75cm}}
     \hline
     \thickhline
      \bf Benchmark   &  \bf Easy pair & \bf Hard pair \\
       \hline
     \thickhline
     Newsroom & ('abstractive','lede3') & ('abstractive','fragments') \\
     & ('abstractive','textrank') & ('pointer\_n','pointer\_s') \\
     & ('fragments','lede3') & ('textrank','lede3') \\
     & ('fragments','textrank') & ('pointer\_c','textrank') \\
     & ('abstractive','pointer\_s') & ('pointer\_s','lede3')\\
     & ('fragments','pointer\_s') & ('pointer\_n','textrank') \\
         \hline
     \thickhline
    \end{tabular}
    }
    \caption{System pairs in TextSumm (Newsroom).}
    \label{tab:syspairs-summ3}
    
\end{table}

\begin{table}[!ht]
    \centering
    \resizebox{.45\textwidth}{!}{
    \begin{tabular}{l p{3.75cm} p{3.75cm}}
     \hline
     \thickhline
      \bf Benchmark   &  \bf Easy pair & \bf Hard pair \\
       \hline
     \thickhline
     UniEval-TC & ('Nucleus Decoding (p = 0.5)',
    'New Human Generated') & ('Original Ground Truth', 'New Human Generated')  \\
     & ('Nucleus Decoding (p = 0.5)', 'Original Ground Truth') & ('Nucleus Decoding (p = 0.5)', 'Nucleus Decoding (p = 0.7)') \\
     & ('Nucleus Decoding (p = 0.3)', 'New Human Generated') &  \\
     & ('Nucleus Decoding (p = 0.3)', 'Original Ground Truth') & \\
     & ('Nucleus Decoding (p = 0.7)', 'New Human Generated') & \\
     &('Nucleus Decoding (p = 0.7)', 'Original Ground Truth') & \\
     && \\
     && \\
     \hline
     \thickhline
    \end{tabular}
    }
    \caption{System pairs in DiagGen (UniEval-TC).}
    \label{tab:syspairs-diaggen1}
\end{table}

\begin{table}[!ht]
    \centering
    \resizebox{.45\textwidth}{!}{
    \begin{tabular}{l p{3.75cm} p{3.75cm}}
     \hline
     \thickhline
      \bf Benchmark   &  \bf Easy pair & \bf Hard pair \\
       \hline
     \thickhline
     USR-TC & ('Nucleus Decoding (p = 0.5)',
    'New Human Generated') & ('Original Ground Truth', 'New Human Generated') \\
     & ('Nucleus Decoding (p = 0.5)', 'Original Ground Truth') &  ('Nucleus Decoding (p = 0.5)', 'Nucleus Decoding (p = 0.7)')\\
     & ('Nucleus Decoding (p = 0.3)', 'New Human Generated') & \\
     & ('Nucleus Decoding (p = 0.3)', 'Original Ground Truth') & \\
     & ('Nucleus Decoding (p = 0.7)', 'New Human Generated') & \\
     & ('Nucleus Decoding (p = 0.7)', 'Original Ground Truth') & \\
     \hline
     \thickhline
    \end{tabular}
    }
    \caption{System pairs in DiagGen (USR-TC).}
    \label{tab:syspairs-diaggen2}
\end{table}

\begin{table}[!ht]
    \centering
    \resizebox{.45\textwidth}{!}{
    \begin{tabular}{l p{3.75cm} p{3.75cm}}
     \hline
     \thickhline
      \bf Benchmark   &  \bf Easy pair & \bf Hard pair \\
       \hline
     \thickhline
     USR-PC & ('Seq2Seq', 'New Human Generated') & ('Original Ground Truth', 'New Human Generated') \\
     & ('Seq2Seq', 'Original Ground Truth') & ('KV-MemNN', 'Seq2Seq') \\
     & ('KV-MemNN', 'New Human Generated') & \\
     & ('KV-MemNN', 'Original Ground Truth') & \\
     & ('Language Model', 'New Human Generated') & \\
     & ('Language Model', 'Original Ground Truth') & \\
         \hline
     \thickhline
    \end{tabular}
    }
    \caption{System pairs in DiagGen (USR-PC).}
    \label{tab:syspairs-diaggen3}
\end{table}

\begin{table*}[!ht]
    \centering
    \resizebox{.99\textwidth}{!}{
    \begin{tabular}{l l c c c c c c c c}
     \hline
     \thickhline
    \bf Data & \bf  Difficulty   & \bf  Human & \bf  Perplexity & \bf  BLEU & \bf ROUGE & \bf BERTScore & \bf Single-CTC & \bf Single-CtrlEval & \bf Multi-UniEval \\
       \hline
     \thickhline
       UniEval-summ & Easy &  0.535 & 0.356 & 0.532 & 0.367 & 0.508 & 0.513 & 0.296 & \bf 0.596 \\
       & Hard & 0.145 & 0.295 & \bf 0.325 & 0.155 & 0.306 & 0.296 & 0.232 & 0.269 \\
       summEval & Easy & 0.441 & 0.403 & 0.365 & 0.324 & 0.344 & 0.479 & 0.199 & \bf 0.6 \\
        & Hard & 0.100 & \bf 0.266 & 0.188 & 0.173 & 0.159 & 0.257 & 0.180 & 0.262 \\
       Newsroom & Easy & 0.396 & 0.333 & \bf 0.808 & 0.506 & 0.700 & 0.596 & 0.553 & 0.584 \\
        & Hard & 0.163 & 0.286 & 0.527 & 0.278 & 0.478 & 0.383 & 0.358 & \bf 0.528 \\
        \hline
     \thickhline
    \end{tabular}
    }
    \caption{Kolmogorov-Smirnov (KS) Scores on system-level performance in TextSumm.}
    \label{tab:ks-textsumm}
\end{table*}

\begin{table*}[!ht]
    \centering
    \resizebox{.99\textwidth}{!}{
    \begin{tabular}{l l c c c c c c c c}
     \hline
     \thickhline
    \bf Data & \bf  Difficulty   & \bf  Human & \bf  Perplexity & \bf  BLEU & \bf ROUGE & \bf BERTScore & \bf Single-CTC & \bf Single-CtrlEval & \bf Multi-UniEval \\
       \hline
     \thickhline
       UniEval-TC & Easy & \bf 0.686 & 0.283 & 0.194 & 0.303 & 0.261 & 0.375 & 0.144 & \bf 0.565 \\ 
       & Hard & 0.203 & 0.225 & 0.158 & 0.200 & 0.133 & 0.226 & 0.125 & \bf 0.317 \\
       USR-TC & Easy & \bf 0.562 & 0.336 & 0.194 & 0.303 & 0.253 & 0.416 & 0.197 & \bf 0.486 \\
        & Hard & 0.121 & 0.242 & 0.158 & 0.200 & 0.125 & 0.232 & 0.144 & \bf 0.283 \\
       USR-PC & Easy & 0.347 & 0.394 & 0.236 & 0.300 & 0.353 & \bf 0.481 & 0.144 & 0.386 \\
        & Hard & 0.156 & \bf 0.433 & 0.258 & 0.375 & 0.275 & 0.390 & 0.147 & 0.218 \\
        \hline
     \thickhline
    \end{tabular}
    }
    \caption{Kolmogorov-Smirnov (KS) Scores on system-level performance in DiagGen.}
    \label{tab:ks-diaggen}
\end{table*}

\begin{table*}[!ht]
    \centering
    \resizebox{.99\textwidth}{!}{
    \begin{tabular}{l l c  c c c c c}
     \hline
     \thickhline
    \bf Data & \bf  Difficulty   & \bf  Human & \bf  Perplexity & \bf BERTScore & \bf Single-CTC & \bf Single-CtrlEval & \bf Multi-UniEval \\
       \hline
     \thickhline
       UBER-Topic & Easy & 0.213 & 0.316 & 0.132 & 0.173 &  0.144 & 0.025  \\ 
       & Hard & 0.048  & 0.134 & 0.105 & 0.074 & 0.073 & 0.027 \\
       CTRL-Topic & Easy & 0.106 & 0.101 & 0.304 &  0.165 & 0.249 & 0.136 \\
        & Hard & 0.079 &  0.113 & 0.097 & 0.075 & 0.092 &  0.096 \\
        \hline
     \thickhline
    \end{tabular}
    }
    \caption{Kolmogorov-Smirnov (KS) Scores on system-level performance in CtrlGen.}
    \label{tab:ks-ctrlgen}
\end{table*}

\begin{table*}[!ht]
    \centering
    \resizebox{.99\textwidth}{!}{
    \begin{tabular}{l l  c c c c c c c}
     \hline
     \thickhline
    \bf Data & \bf  Difficulty    & \bf  Perplexity & \bf  BLEU & \bf ROUGE & \bf BERTScore & \bf Single-CTC & \bf Single-CtrlEval & \bf Multi-UniEval \\
       \hline
     \thickhline
       UniEval-summ & Easy &  0.711 & 0.933 & \bf 0.989 & \bf 0.989  & 0.924  & 0.622 & \bf 0.989 \\
       & Hard & 0.648  & 0.758 & 0.612 & 0.709 & 0.688  & 0.685 & \bf 0.803 \\
       summEval & Easy &  0.752  & 0.919 & 0.776 & 0.943 & 0.943 & 0.752  & \bf 0.983 \\
        & Hard & 0.707  & 0.647  & 0.673 & 0.613 & \bf 0.762 & 0.693  & 0.730 \\
       Newsroom & Easy & 0.444 & \bf 1.000 & 0.889 & \bf 1.000 & 0.963  & \bf 1.000 & 0.833 \\
        & Hard & 0.555  & \bf 0.889  & \bf 0.889 & \bf 0.889 & 0.870 & \bf 0.889 & 0.833 \\
        \hline
     \thickhline
    \end{tabular}
    }
    \caption{Preference similarity in TextSumm.}
    \label{tab:pref-sim-textsumm}
\end{table*}

\begin{table*}[!ht]
    \centering
    \resizebox{.99\textwidth}{!}{
    \begin{tabular}{l l  c c c c c c c}
     \hline
     \thickhline
    \bf Data & \bf  Difficulty    & \bf  Perplexity & \bf  BLEU & \bf ROUGE & \bf BERTScore & \bf Single-CTC & \bf Single-CtrlEval & \bf Multi-UniEval \\
       \hline
     \thickhline
       UniEval-summ & Easy & 0.889  & \bf 1.000 & \bf 1.000 & 0.667 & \bf 1.000 & 0.444 & 1.000 \\
       & Hard & 0.611  & 0.722 & \bf 0.944 & 0.833 & 0.722 & 0.388 & 0.722 \\
       summEval & Easy & 0.778  & \bf 1.000 & \bf 1.000 & 0.667 & \bf 1.000 & 0.629 & 0.926 \\
        & Hard &  0.500 & 0.833  & \bf 0.944 & 0.833  & 0.722 & 0.593 & 0.796 \\
       Newsroom & Easy & \bf 1.000 & \bf 1.000 & 0.778  & 0.667 & \bf 1.000 & 0.741 & 0.944 \\
        & Hard &  0.611 & 0.722  & 0.833  & 0.667 & 0.833  & \bf 0.889  & 0.833  \\
        \hline
     \thickhline
    \end{tabular}
    }
    \caption{Preference similarity in DiagGen.}
    \label{tab:pref-sim-diaggen}
\end{table*}

\begin{table*}[!ht]
    \centering
    \resizebox{.99\textwidth}{!}{
    \begin{tabular}{l l  c c c c c}
     \hline
     \thickhline
    \bf Data & \bf  Difficulty    & \bf  Perplexity  & \bf BERTScore & \bf Single-CTC & \bf Single-CtrlEval & \bf Multi-UniEval \\
       \hline
     \thickhline
       UBER-Topic & Easy &  0.667   & 0.667 & 0.667   & 0.667  & 0.667  \\
       & Hard & 0.333   & \bf 1.000 & 0.778   & 0.555 & 0.417 \\
       CTRL-Topic & Easy &  0.333   & \bf 1.000 & 0.611 & 0.555 & 0.417  \\
        & Hard &  0.333  & \bf 1.000 &  0.666  & 0.555 & 0.333 \\
        \hline
     \thickhline
    \end{tabular}
    }
    \caption{Preference similarity in CtrlGen.}
    \label{tab:pref-sim-ctrlgen}
\end{table*}

\begin{figure*}[!ht]
    \centering
       \begin{subfigure}[t]{.48\textwidth}
         \centering
         \includegraphics[width=\linewidth]{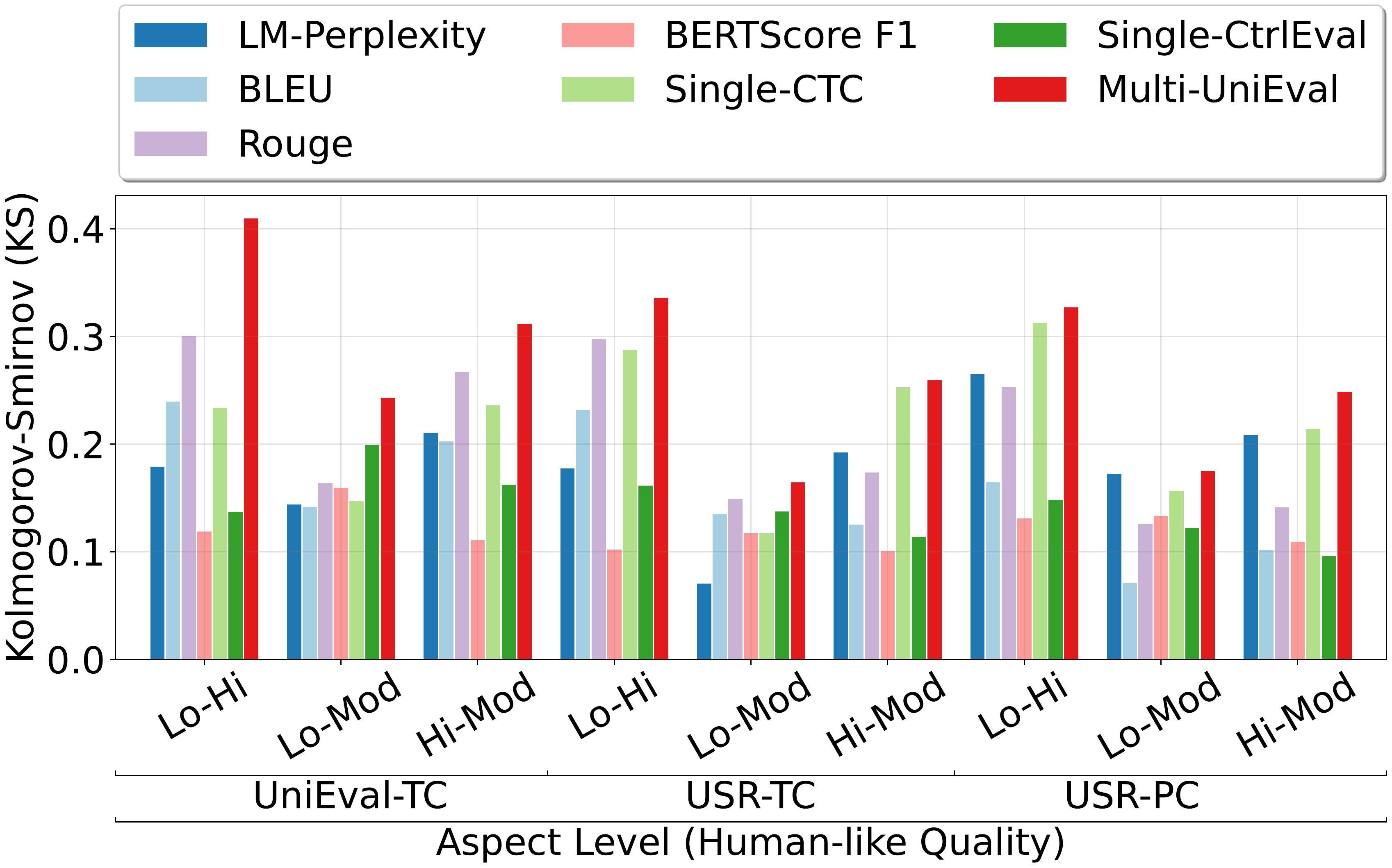}
         \caption{Identifying different levels of quality.}
         \label{}
       \end{subfigure}
    \hfill
    \begin{subfigure}[t]{.48\textwidth}
         \centering
         \includegraphics[width=\linewidth]{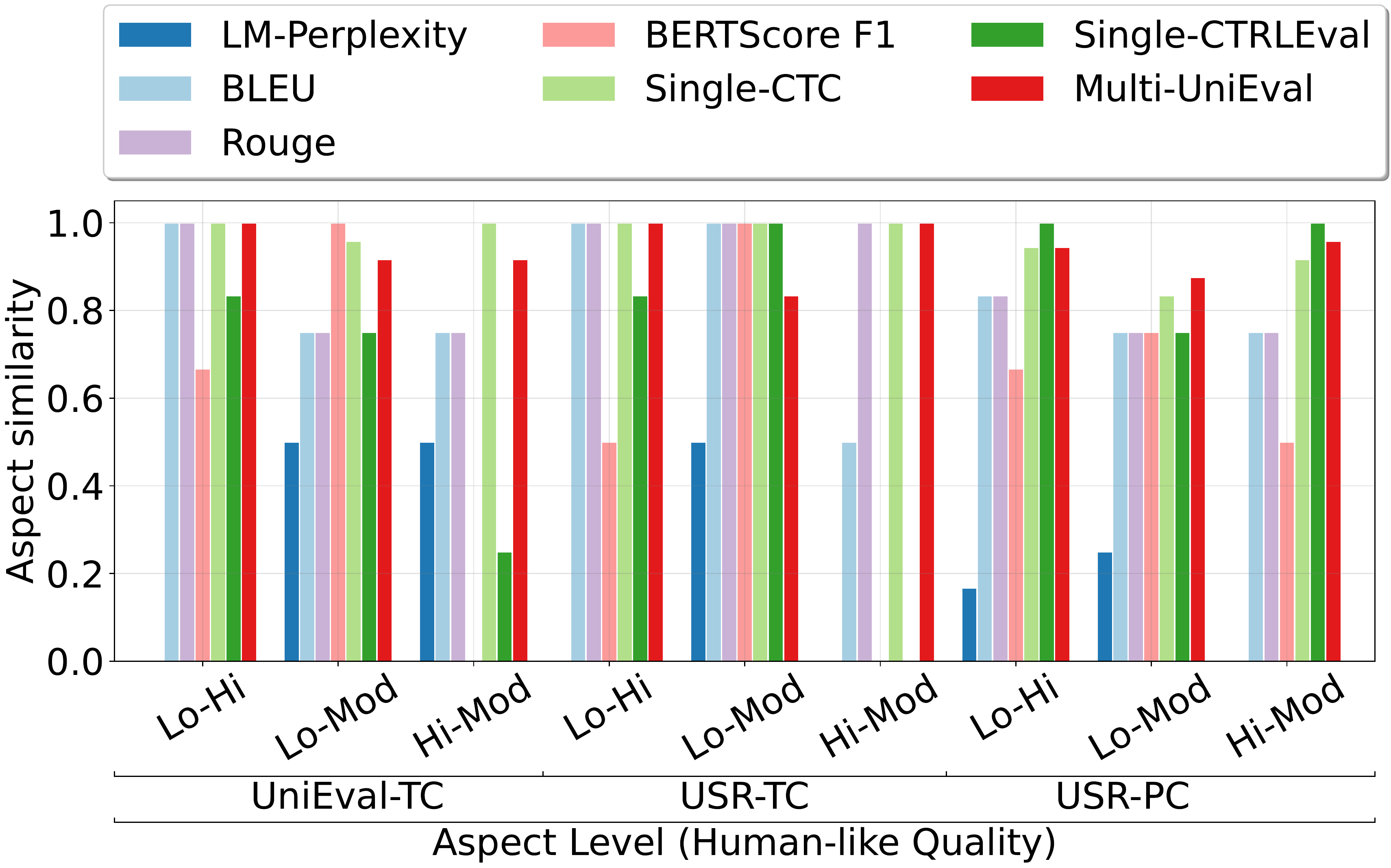}
         \caption{Rank/Preference similarity to human.}
         \label{}
       \end{subfigure}
    \caption{Aspect-level evaluation in Dialogue Response Generation (DiagGen). \textbf{Left}: Kolmogorov-Smirnov (KS) score on discerning between two different levels of human-like quality -- Higher is better $[0,1]$. \textbf{Right}: Similarity to the rank of the aspect-levels based on human scores -- Higher is better $[0,1]$. Lo-Hi: Low vs. High quality (e.g. Poor Coherent vs. Highly coherent), Lo-Mod: Low vs. Moderate. Hi-Mod: High vs. Moderate.
    }
    \label{fig:ks-diag-aspect}
\end{figure*}

\end{document}